\documentclass{article}

\usepackage[main, final]{neurips_2026}

\usepackage[utf8]{inputenc}
\usepackage[T1]{fontenc}
\usepackage[colorlinks=true,allcolors=teal]{hyperref}
\usepackage{url}
\usepackage{booktabs}
\usepackage{amsfonts}
\usepackage{amsmath, amssymb, amsthm, mathtools}
\usepackage{microtype}
\usepackage{xcolor}
\usepackage{enumitem}
\usepackage{graphicx}
\usepackage{caption}
\usepackage{subcaption}
\usepackage{float}
\usepackage{wrapfig}
\usepackage{multirow}
\usepackage{tikz}
\usetikzlibrary{arrows.meta, positioning, calc, patterns, decorations.pathreplacing}

\title{Embeddings for Preferences, Not Semantics\thanks{Code: \url{https://github.com/cartgr/Embeddings-for-Preferences}\\\hspace*{4.0ex}Model: \url{https://huggingface.co/cartgr/embeddings-for-preferences-st5-xl}}}

\author{%
  Carter Blair \\
  Harvard University \\
  \texttt{carterblair@g.harvard.edu} \\
  \And
  Ariel D.\ Procaccia \\
  Harvard University \\
  \texttt{arielpro@g.harvard.edu} \\
  \And
  Milind Tambe \\
  Harvard University \\
  \texttt{tambe@g.harvard.edu} \\
}

\newtheorem{theorem}{Theorem}

\newcommand{\X}{\mathcal{X}}
\newcommand{\E}{\mathbb{E}}

\newcommand{\ip}[2]{\left\langle #1, #2 \right\rangle}

\begin{document}

\maketitle

\begin{abstract}
Modern AI is opening the door to collective decision-making in which participants express their views as free-form text rather than voting on a fixed set of candidates. A natural idea is to embed these opinions in a vector space so that the substantial literature on facility location problems and fair clustering can be brought to bear. But standard text embeddings measure semantic similarity, whereas distances in facility location problems and fair clustering require what we call \textit{preferential similarity}: a participant's agreement with a piece of text should be inversely related to their distance from it. Off-the-shelf embeddings inherit a coarse preference signal through a correlation between semantic and preferential similarity, but fail to capture preferences when the correlation breaks. We formalize this as an invariance problem: text embedding models encode both a preference-relevant signal (stance and values) and semantic nuisance (style and wording), and the two are observationally correlated, so a geometry that relies on nuisance can appear preference-correct even when it is not. We show that synthetic training data designed to break this correlation provably shifts the optimal scorer away from nuisance-dominated cosine and significantly improves preference prediction across 11 online deliberation datasets.
\end{abstract}

\section{Introduction}

Many emerging systems for collective decision-making allow participants to express their preferences in free-form text instead of voting on predetermined candidate sets. For example, two influential online deliberation platforms, Polis and Remesh, allow participants to write statements and vote on statements written by others \citep{small2021polis}. Similarly, in generative social choice, participants express their preferences using free-form text, which are then aggregated into a slate of representative statements \citep{fish2026generative}. A commonality among these systems is that they require methods to group participants and to estimate a participant's utility for statements they did not vote on. Since inputs are free-form text, a natural idea is to embed each participant's text using a text embedding model. This would permit the grouping of participants via clustering in the embedding space and distances could be used to estimate participants' utility for statements they did not vote on. Further, it would allow for novel applications of ideas from the facility location and fair clustering literature \citep{feldman2016voting, chen2019proportionally, micha2020proportionally, kellerhals2024proportional}.

However, off-the-shelf text embedding models are mainly trained and evaluated on semantic tasks such as retrieval, textual similarity, and natural-language inference \citep{muennighoff-etal-2023-mteb}. These tasks tend to reward placing texts close together in embedding space when they discuss the same topic or answer a similar query. They do not necessarily require close points to be mutually endorsable or, in other words, preferentially similar.

The distinction between semantic and preferential similarity is important to understand if embeddings are to be used inside preference aggregation procedures. To make the importance clear, it is useful to imagine two statements from either side of a controversial political debate. They may share the same style and topic, as well as many of the same words, all of which a generic text embedding model may pick up on. However, they would not be mutually endorsed. Table~\ref{tab:example} gives an example where a small alteration of a statement produces near-identical surface similarity to the anchor yet significantly alters the preferential similarity. A standard embedding model (ST5-XL) scores the altered distractor statement as more similar than an opinion-aligned alternative expressed in different words.

\begin{table*}[t]
\centering
\small
\caption{
A hard triplet illustrating the mismatch between semantic and preferential similarity. The semantic distractor uses the anchor's wording but reverses its stance, while the preference match preserves the stance but changes the wording. A base embedding model ranks the distractor closer to the anchor. A preference-tuned and topic-specific projected embedding produce the correct ranking.
}
\label{tab:example}
\begin{tabular*}{\textwidth}{@{\extracolsep{\fill}}p{1.3cm}p{6.8cm}ccc@{}}
\toprule
 & & \multicolumn{3}{c}{\textbf{Similarity}} \\
\cmidrule(lr){3-5}
\textbf{Role} & \textbf{Text} & \textbf{Base} & \textbf{Tuned (\S\ref{sec:method})} & \textbf{Projected (\S\ref{sec:probes})} \\
\midrule
Anchor & ``It should be \textbf{legal}. Religion has \textbf{no} place in politics if we truly want to be free.'' & & & \\
\addlinespace
Distractor & ``It should be \textbf{illegal}. Religion has \textbf{a} place in politics if we truly want to be free.'' & \textbf{0.87} & 0.79 & $-0.74$ \\
\addlinespace
Match & ``To ensure true freedom, it's crucial that legislation remains secular, without religious influence.'' & 0.82 & \textbf{0.81} & \textbf{0.99} \\
\bottomrule
\end{tabular*}
\end{table*}

We frame the mismatch between semantic and preferential geometry as an invariance problem \citep{achille2018emergence}: a preference geometry should be invariant to wording and style, and sensitive only to stance and values. Generic embedding models do not have this invariance, because they encode topical and stylistic features that are useful for retrieval and similarity tasks but unrelated to whether two statements would be endorsed by the same person. On natural deliberation data, this gap is partly hidden because semantic and preferential similarity are correlated: people who share a stance often share wording. In \S\ref{sec:framework} we make the confounding explicit by decomposing the cosine margin into a preference component and a nuisance component. Cosine weights the two equally, so when they agree (as on natural data) it looks correct, but when they disagree (as on the hard triplet in Table~\ref{tab:example}) the nuisance dominates and cosine fails.

Our method follows from this diagnosis. We synthesize triplets consisting of an anchor, a preference match with different wording, and a semantic distractor with high surface overlap but opposite stance. Training on these triplets forces cosine to down-weight nuisance variation that ordinary data leaves confounded with preference. We prove that, under this hard-triplet distribution, the Bradley-Terry risk is strictly decreased by reducing the nuisance contribution below cosine's unit weighting. Empirically, our method significantly improves performance on hard triplets and on preference prediction across 11 online deliberation datasets. 

When per-topic votes are available, which is common in online deliberation platforms, the same framework suggests an even simpler method: learning a low-rank projection of the frozen embedding. Despite its simplicity, this projected embedding performs very well in practice.

To summarize, our contributions are as follows:
\begin{enumerate}[noitemsep, topsep=0pt, leftmargin=*]
    \item We diagnose the mismatch between semantic and preferential similarity as an invariance problem, and formalize it via a decomposition of the cosine margin into preference signal and nuisance.
    \item We introduce a hard-triplet synthesis procedure and prove that training on it strictly decreases the Bradley-Terry risk relative to standard cosine.
    \item We demonstrate substantial gains on hard triplets and on preference prediction across 11 online deliberation datasets.
    \item We show that when per-topic votes are available, a low-rank projection of frozen embeddings outperforms full preference tuning.
    
\end{enumerate}

\section{Related Work}
\label{sec:related}
A growing body of work on collective decision-making over free-form text relies on some geometry over participants or statements: Polis derives opinion maps from vote matrices \citep{small2021polis}, generative social choice groups statements in an LLM-defined feature space to produce representative slates \citep{fish2026generative}, \cite{blair2025approxconsensus} model consensus as a region of embedding space, and \cite{de2026question} use cosine similarity of embeddings as participant utility when auditing justified representation in slates of questions. Our goal in this paper is to answer a prerequisite question. Namely, do distances in general-purpose sentence embedding spaces reflect preferential similarity? And, if not, can they be realigned so that they do?

Another line of work fine-tunes sentence encoders for opinion-related tasks, including stance-aware embeddings for opinion mining \citep{ghafouri-etal-2024-love} and sparsity-aware embeddings for contradiction retrieval \citep{xu2024sparsecl}. However, in \S\ref{sec:results} we find that neither works well for our task. Our recipe is also related to SimCSE \citep{gao2021simcse}, which uses NLI entailments as positives and contradictions as hard negatives. Our task is different and we go a step further by engineering triplets where the nuisance signal intentionally points in the wrong direction (see Appendix \ref{app:normal-triplet} for the relevant ablation). An extended related work discussion is in Appendix~\ref{app:related}.
\section{Evaluation Setup}
\label{sec:setup}

Before presenting our diagnosis and method, we describe the evaluation data and metrics used throughout. We evaluate on 11 datasets from three deliberation platforms. Generative social choice (GSC) \citep{fish2026generative} provides surveys in which participants write free-text opinions and then rate AI-generated statements (two abortion surveys and one on chatbot personalization). Remesh provides binary agree/disagree votes on others' open-ended responses across three topics (campus protests, foreign intervention, right to assemble). Polis \citep{small2021polis} provides comment-level agree/disagree votes across five conversations (Seattle minimum wage, Bowling Green, Brexit, Canadian electoral reform, universal basic income). Some participants author comments in addition to voting. Together these span text lengths from short Polis comments to multi-paragraph GSC opinions and cover 1{,}462 participants, 3{,}958 statements, and 1.46M pairwise preference triplets. Dataset details and URLs are in Appendix~\ref{app:datasets}.

For each participant, we construct preference triplets $(a, p, n)$ where $a$ is the participant's own written text (the anchor), $p$ is a statement they rated more favorably, and $n$ is one they rated less favorably. Candidate models score each triplet by computing a similarity margin $s(a, p) - s(a, n)$; the triplet is correct if this margin is positive. We call the fraction of correctly ordered triplets \emph{triplet accuracy}, and \emph{cosine accuracy} when the scorer $s$ is cosine similarity. In \S\ref{sec:probes} we also evaluate the ideal-point scorer, for which $s$ is the learned distance-based utility. Training details for our methods are given in \S\ref{sec:recipe}, \S\ref{sec:probes}, and Appendix~\ref{app:additional}.

\section{Diagnosing Embedding Models}
\label{sec:diagnosis}

We begin by characterizing the preference signal in existing embedding models. We introduce a formal framework (\S\ref{sec:framework}), show that cosine is an approximation to the ideal-point utility margin, present empirical evidence for a correlation between nuisance and preference signal in the natural-data regime (\S\ref{sec:bands}), and diagnose the hard-triplet failure (\S\ref{sec:hard}).

\subsection{Formal framework}
\label{sec:framework}

Let $\psi : \X \to \mathbb{R}^d$ denote a pretrained encoder producing unit-norm embeddings. Within the embedding, preference on a given topic is governed by a \emph{preference subspace} $S \subseteq \mathbb{R}^d$ of dimension $k \ll d$ that carries the stance-relevant structure. Let $P_S$ denote orthogonal projection onto $S$ and $P_{S^\perp}$ onto its complement, and abbreviate $\psi_S := P_S \psi$, $\psi_\perp := P_{S^\perp} \psi$. Each participant $v$ writes anchor text $a_v$ and has \emph{ideal point} $u_v := \psi_S(a_v) \in S$, which is the projection of their own embedding onto $S$. Different participants have different anchors and thus different ideal points in $S$. For example, pro-choice and pro-life users live in opposite regions along the stance axis while sharing $S$ as the relevant direction of variation.

We model the utility of candidate statement $j$ as the negative squared Euclidean distance to the participant's ideal point within $S$.\footnote{The derivations below extend to the general Mahalanobis case $-\|\psi_S(a_v) - \psi_S(x_j)\|_M^2$ for any PSD $M$ on $S$: absorbing $\sqrt{M}$ into the encoder reduces it to the Euclidean case with no other change.} 
Expanding the square,
\[
  U^*(v, j) \;=\; -\|\psi_S(a_v) - \psi_S(x_j)\|^2
            \;=\; 2\ip{\psi_S(a_v)}{\psi_S(x_j)}
                  - \|\psi_S(a_v)\|^2
                  - \|\psi_S(x_j)\|^2.
\]
Pairwise preferences follow Bradley-Terry: $\Pr[p \succ n \mid v] = \bigl(1+e^{-(U^*(v,p) - U^*(v,n))}\bigr)^{-1}$. For within-anchor rankings, the $\|\psi_S(a_v)\|^2$ term is constant across candidates and cancels out, so the ranking-relevant utility margin is
\begin{equation}
\label{eq:util-margin}
  U^*(v,p) - U^*(v,n)
  \;=\; 2\underbrace{\ip{\psi_S(a_v)}{\psi_S(x_p) - \psi_S(x_n)}}_{\Delta_S}
       \;+\; \underbrace{\|\psi_S(x_n)\|^2 - \|\psi_S(x_p)\|^2}_{\Delta_{\text{norm}}}.
\end{equation}
The first term $\Delta_S$ is bilinear in the anchor and item projections and the second $\Delta_{\text{norm}}$ is an item-specific quadratic that depends only on candidate norms within $S$.

\paragraph{Cosine as an approximation.} A cosine scorer on an off-the-shelf embedding model does not know $S$ and uses all $d$ dimensions equally. Because $P_S$ and $P_{S^\perp}$ map into orthogonal subspaces, the cosine margin decomposes additively as
\begin{equation}
\label{eq:decomp}
  s(a_v,p) - s(a_v,n)
  \;=\; \underbrace{\ip{\psi_S(a_v)}{\psi_S(x_p) - \psi_S(x_n)}}_{\Delta_S}
       \;+\; \underbrace{\ip{\psi_\perp(a_v)}{\psi_\perp(x_p) - \psi_\perp(x_n)}}_{\Delta_T}.
\end{equation}
Comparing~\eqref{eq:decomp} with~\eqref{eq:util-margin}, cosine captures the bilinear part of the utility margin (up to a factor of two) but misses the item-norm difference $\Delta_{\mathrm{norm}}$ entirely, and contributes an out-of-subspace nuisance $\Delta_T$ that has no utility counterpart. Cosine is therefore an approximation to the ideal-point utility margin, tight when $\Delta_{\mathrm{norm}}$ is small and when $\Delta_T$ aligns with $\Delta_S$. In \S\ref{sec:probes} we find empirical evidence that this model describes real preferences well.

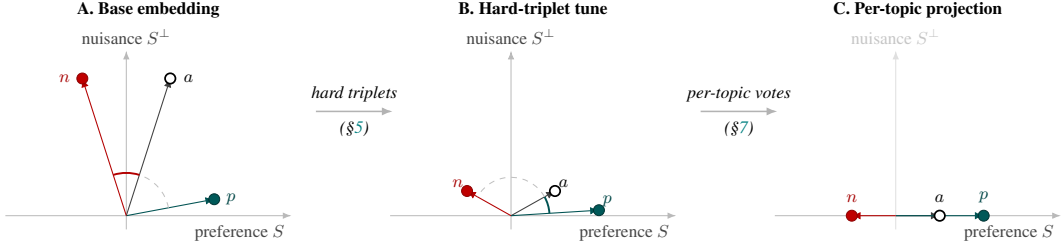
\begin{figure}[t]
\centering
\resizebox{\textwidth}{!}{%
\begin{tikzpicture}[
  anchpt/.style={circle, inner sep=1.9pt, fill=white, draw=black, line width=0.9pt},
  matchpt/.style={circle, inner sep=2.0pt, fill=teal!70!black, draw=teal!40!black, line width=0.4pt},
  distpt/.style={circle, inner sep=2.0pt, fill=red!75!black, draw=red!55!black, line width=0.4pt},
  vec/.style={-{Latex[length=1.6mm, width=1.4mm]}, line width=0.55pt},
  axlabel/.style={font=\small, text=gray!55!black},
  panellabel/.style={font=\bfseries\small},
]

\begin{scope}
  \draw[-{Latex[length=1.5mm]}, gray!55, thin] (-2.2, 0) -- (3.0, 0);
  \node[axlabel, anchor=north east] at (3.0, -0.05) {preference $S$};
  \draw[-{Latex[length=1.5mm]}, gray!55, thin] (0, -0.25) -- (0, 3.0) node[above, axlabel]{nuisance $S^\perp$};

  \coordinate (O) at (0,0);
  \coordinate (A) at (0.8, 2.5);
  \coordinate (P) at (1.6, 0.3);
  \coordinate (N) at (-0.8, 2.5);

  \draw[vec, black!75]      (O) -- (A);
  \draw[vec, teal!65!black] (O) -- (P);
  \draw[vec, red!70!black]  (O) -- (N);

  \draw[red!70!black, line width=1pt]       (72.25:0.78) arc (72.25:107.75:0.78);
  \draw[gray!50, dashed, line width=0.4pt]  (10.6:0.78)  arc (10.6:72.25:0.78);

  \node[anchpt]  at (A) {};
  \node[matchpt] at (P) {};
  \node[distpt]  at (N) {};

  \node[right=1.2mm, font=\small]                at (A) {$a$};
  \node[right=1mm,   font=\small, teal!55!black] at (P) {$p$};
  \node[left=1mm,    font=\small, red!65!black]  at (N) {$n$};

  \node[panellabel] at (0.4, 3.75) {A.\ Base embedding};
\end{scope}

\begin{scope}[xshift=7cm]
  \draw[-{Latex[length=1.5mm]}, gray!55, thin] (-2.2, 0) -- (3.0, 0);
  \node[axlabel, anchor=north east] at (3.0, -0.05) {preference $S$};
  \draw[-{Latex[length=1.5mm]}, gray!55, thin] (0, -0.25) -- (0, 3.0) node[above, axlabel]{nuisance $S^\perp$};

  \coordinate (O) at (0,0);
  \coordinate (A) at (0.8, 0.45);
  \coordinate (P) at (1.6, 0.1);
  \coordinate (N) at (-0.8, 0.45);

  \draw[vec, black!75]      (O) -- (A);
  \draw[vec, teal!65!black] (O) -- (P);
  \draw[vec, red!70!black]  (O) -- (N);

  \draw[teal!65!black, line width=1pt]       (3.58:0.7) arc (3.58:29.4:0.7);
  \draw[gray!50, dashed, line width=0.4pt]   (29.4:0.7)  arc (29.4:150.6:0.7);

  \node[anchpt]  at (A) {};
  \node[matchpt] at (P) {};
  \node[distpt]  at (N) {};

  \node[above right=-0.8mm, font=\small]                 at (A) {$a$};
  \node[above right=-0.8mm, font=\small, teal!55!black]  at (P) {$p$};
  \node[above left=-0.8mm, font=\small, red!65!black]    at (N) {$n$};

  \node[panellabel] at (0.4, 3.75) {B.\ Hard-triplet tune};
\end{scope}

\begin{scope}[xshift=14cm]
  \draw[-{Latex[length=1.5mm]}, gray!55, thin] (-2.2, 0) -- (3.0, 0);
  \node[axlabel, anchor=north east] at (3.0, -0.05) {preference $S$};
  \draw[-{Latex[length=1.5mm]}, gray!25, thin] (0, -0.25) -- (0, 3.0) node[above, axlabel, text=gray!45]{nuisance $S^\perp$};

  \coordinate (O) at (0,0);
  \coordinate (A) at (0.8, 0);
  \coordinate (P) at (1.6, 0);
  \coordinate (N) at (-0.8, 0);

  \draw[vec, black!75]      (O) -- (A);
  \draw[vec, teal!65!black] (O) -- (P);
  \draw[vec, red!70!black]  (O) -- (N);

  \node[anchpt]  at (A) {};
  \node[matchpt] at (P) {};
  \node[distpt]  at (N) {};

  \node[above=1.2mm, font=\small]                at (A) {$a$};
  \node[above=1.2mm, font=\small, teal!55!black] at (P) {$p$};
  \node[above=1.2mm, font=\small, red!65!black]  at (N) {$n$};

  \node[panellabel] at (0.4, 3.75) {C.\ Per-topic projection};
\end{scope}

\draw[-{Latex[length=2mm]}, thick, gray!60]
  (3.45, 1.9) -- node[above=0.5mm, font=\small\itshape, text=black] {hard triplets}
                node[below=0.5mm, font=\small\itshape, text=black] {(\S\ref{sec:method})}
                (4.85, 1.9);
\draw[-{Latex[length=2mm]}, thick, gray!60]
  (10.45, 1.9) -- node[above=0.5mm, font=\small\itshape, text=black] {per-topic votes}
                 node[below=0.5mm, font=\small\itshape, text=black] {(\S\ref{sec:probes})}
                 (11.85, 1.9);
\end{tikzpicture}%
}
\caption{
A hard triplet: anchor $a$, preference-match $p$ (same stance, different wording), and semantic distractor $n$ (opposite stance, same wording) with preference subspace $S$ horizontal and nuisance $S^\perp$ vertical.
(A) In the pretrained embedding, $n$ shares $a$'s nuisance component, so cosine ranks $n$ above $p$.
(B) Fine-tuning on counterfactual hard triplets downweights $\psi_\perp$ (Theorem~\ref{thm:hardtriplet}).
(C) With per-topic labels, a rank-$r$ projection $L^\top$ maps onto $S$ directly, discarding $S^\perp$.}
\label{fig:geometry}
\end{figure}

Figure~\ref{fig:geometry} illustrates the three stages in this paper: the diagnosis that base cosine overweights the nuisance direction (\S\ref{sec:diagnosis}), hard-triplet fine-tuning which suppresses $\Delta_T$ (\S\ref{sec:method}), and a per-topic rank-$r$ projection which projects onto $S$ directly (\S\ref{sec:probes}).

\subsection{Proximity bands: cosine carries a preference signal}
\label{sec:bands}

With the framework in place, we first ask what cosine similarity on base
embedding models captures on natural deliberation data. This diagnostic is
pairwise (between an anchor statement and a candidate statement) rather than triplet-based. For each participant, we compute cosine similarity from their anchor embedding to every statement they voted on, bin these pairs into quintiles by similarity, and compute the approval rate per band. We use the binary-vote datasets (Remesh and Polis), where approval is directly observed.

\begin{wrapfigure}{r}{0.42\textwidth}
\vspace{-0.5em}
\centering
\includegraphics[width=\linewidth]{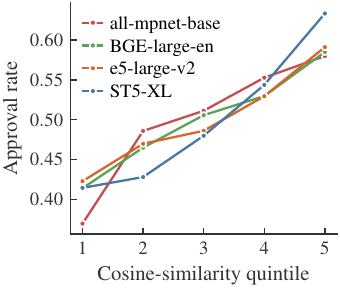}
\caption{Approval rate by cosine-similarity quintile, pooled across the
binary-vote evaluation datasets.}
\label{fig:bands}
\vspace{-2.5em}
\end{wrapfigure}

Approval rises monotonically across all four encoders (Figure~\ref{fig:bands}). The spread
between the most distant and most similar quintile is roughly between 15 and 20 percentage
points for each encoder. From this it is clear that cosine carries some preference signal, as statements closer to a participant's own (according to cosine similarity) are more likely to be approved by that participant.

However, where this signal comes from is less clear. A high cosine score can arise because the candidate is close to the anchor in the preference subspace, or because it shares topic, wording, style, or affect that happen to correlate with preference in naturally occurring text on deliberation platforms. More specifically, at the pair level similarity can be decomposed as
\[
  s(a_v,x_j)
  =
  \langle \psi(a_v), \psi(x_j) \rangle
  =
  \underbrace{\langle \psi_S(a_v), \psi_S(x_j) \rangle}_{s_S(a_v,x_j)}
  +
  \underbrace{\langle \psi_\perp(a_v), \psi_\perp(x_j) \rangle}_{s_T(a_v,x_j)}.
\]
Figure \ref{fig:bands} shows that the sum is predictive of approval but it does not identify whether the cause is the preference term $s_S$, the nuisance term $s_T$, or the two terms moving together. On natural deliberation data it could be the case that people who share a stance often also share wording and style. If this were the case, semantic similarity and preferential similarity would be observationally correlated.

The same ambiguity appears in the triplet margin used throughout the paper. For
a triplet $(a,p,n)$ with $p$ preferred to $n$, differencing the pairwise scores
gives
\[
  s(a,p)-s(a,n)=\Delta_S+\Delta_T.
\]

Cosine gives $\Delta_S$ and $\Delta_T$ unit weight, but their effective strength on a dataset can be very different. The contribution of each term depends not only on the scorer's coefficient, but also on the norm, variance, dimensionality, and label-alignment of the component in the observed triplets. If $\Delta_S$ is large and reliably aligned with approval, then cosine succeeds because the embedding already contains a strong preference geometry. If $\Delta_T$ is large and merely correlated with approval, then cosine can also succeed on natural data, but for the wrong reason --- namely, by using semantic features that happen to track preferences in the observed distribution.

The proximity bands therefore leave open two explanations. Base embeddings may genuinely put nearby points near each other because they are preferentially similar. Alternatively, base embeddings may mostly capture semantic similarity, which may appear preference-aware only because semantic similarity is correlated with preferential similarity in data from online deliberation platforms. The next diagnostic is designed to ascertain which of the two explanations is correct by putting the two sources of signal in conflict.

\subsection{Hard triplets: separating preference from surface similarity}
\label{sec:hard}

To test whether the apparent preference signal in base embedding models results from a correlated nuisance component or from a true preference awareness, we construct hard triplets that deliberately decouple preference alignment from surface similarity. For each anchor, we generate two comparison statements: a \emph{semantic distractor} that shares the anchor's vocabulary but takes the opposite stance, and a \emph{preference match} that shares the anchor's opinion but uses different vocabulary. A model that relies on surface features will score the distractor higher, whereas a model that captures preferential similarity will score the match higher.

We generate 875 such triplets (100 per evaluation dataset where available) by rewriting real participant anchors with GPT-4o (prompt in Appendix~\ref{app:prompts}). On these triplets, cosine fails across every encoder family we test. Base sentence-T5-XL reaches only 48.3\%, near chance. Three widely-used encoders perform far worse: e5-large-v2 (26.7\%), BGE-large (6.3\%), and all-mpnet-base (8.2\%). These three rank the semantic distractor above the preference match 70–95\% of the time. Meanwhile these same encoders score 58--60\% on the natural-data triplets (Table~\ref{tab:main-hard}). The drops, of up to roughly 50 percentage points, from natural to hard data suggest that cosine's apparent preference signal on natural data comes almost entirely from a nuisance component that is correlated with preferences on natural data.

\begin{table}[ht]
\centering
\small
\caption{Triplet accuracy (\%) on the natural datasets (test split) vs. the hard triplets described in \S\ref{sec:hard}.}
\label{tab:main-hard}
\begin{tabular*}{\textwidth}{@{\extracolsep{\fill}}lccc@{}}
\toprule
\textbf{Encoder} & \textbf{Natural (mean, 11 ds.)} & \textbf{Hard ($n{=}875$)} & $\Delta$ \\
\midrule
sentence-T5-XL        & 65.2 & 48.3 & -16.9 \\
e5-large-v2           & 60.4 & 26.7 & -33.7 \\
BGE-large-en-v1.5     & 59.2 & \phantom{0}6.3 & -52.9 \\
all-mpnet-base-v2     & 57.8 & \phantom{0}8.2 & -49.6 \\
\bottomrule
\end{tabular*}
\end{table}

In sum, hard triplets show that the natural-data signal from cosine on base embedding models is not robust when surface similarity and preference alignment conflict. This suggests using such conflicts not just as a diagnostic, but as supervision.

\section{Decorrelated Preference Tuning}
\label{sec:method}

The hard-triplet diagnostic suggests a simple intervention: train on cases where surface similarity and preference alignment disagree. However, \S\ref{sec:bands} and \S\ref{sec:hard} suggest that these cases are rare in natural data. Therefore, we construct synthetic counterfactual triplets that break this correlation; we tune the encoder on them so that cosine learns to follow the preference-aligned item rather than the distractor that is similar in a superficial sense. We call this procedure \textit{decorrelated preference tuning} (DPT).

\subsection{Why hard triplets are corrective}
\label{sec:train-out}

We first formalize the direction in which hard-triplet training should move a scorer. Consider the class of bilinear scorers
\[
s_A(a,x) = \psi(a)^\top A\psi(x),
\]
with symmetric $A\in\mathbb{R}^{d\times d}$. To make the nuisance weight explicit, this can be written as
\[
A(B,\lambda) = B + \lambda P_{{S^\perp}},
\]

where $B = P_SBP_S$ only acts on the preference subspace and $\lambda$ scales the nuisance. This family includes cosine on unit-norm embeddings when $B=P_S$ and $\lambda=1$, and the $S$-projected scorer when $B=P_S$ and $\lambda=0$. 

For a triplet $(a,p,n)$, the margin is
\[
\psi(a)^\top A(B,\lambda)(\psi(p)-\psi(n)) = \Delta_B + \lambda\Delta_T,
\]
where $\Delta_B := \psi_S(a)^\top B(\psi_S(p)-\psi_S(n))$ and where $\Delta_T=\langle \psi_\perp(a),\psi_\perp(p)-\psi_\perp(n)\rangle$ is the nuisance margin. Thus, $\lambda$ controls how much the scorer relies on the nuisance part of cosine.

Let $\mathcal{G}=\sigma(\psi_S(a),\psi_S(p),\psi_S(n))$ denote the $\sigma$-algebra generated by the in-subspace parts of the triplet. Conditioning on $\mathcal{G}$ means holding the preference subspace projections fixed while averaging over the nuisance subspace variation. 

The hard-triplet condition can be formalized as $\E[\Delta_T\mid\mathcal{G}] \leq 0$ almost surely, with strict inequality on a set of positive probability. In words, once the preference-relevant parts of the triplet are fixed, the nuisance margin does not help the preferred item and, on average, it favors the distractor. Since we assume the preferred item in the hard triplet is aligned with the anchor in the preference subspace,\footnote{The following proof of Theorem~\ref{thm:hardtriplet}, however, does not require this.} this condition formalizes the idea that, in hard triplets, the preference and nuisance components point in opposite directions.

For the scorer $A(B, \lambda)$, define the Bradley-Terry population risk as
\[
R(B,\lambda) = \E[\log(1+e^{-\Delta_B-\lambda\Delta_T})].
\]

The following result shows that, on any triplet distribution satisfying the hard-triplet condition, risk is reduced by decreasing the weight on the nuisance. 

\begin{theorem}
\label{thm:hardtriplet}
If $\mathbb{E}[\Delta_T\mid\mathcal{G}]\leq 0$ a.s., with strict inequality on a set of positive probability, then
\[
R(B, \lambda) < R(B,1) \quad \text{for every } \lambda \in [0,1).
\]
\end{theorem}

The result does not require knowing $S$ or directly optimizing $\lambda$. Instead, simply constructing a distribution of triplets where $\Delta_S$ points in the right direction and $\Delta_T$ points in the wrong direction for each triplet almost surely rewards scorers that put relatively more weight on the preference aligned part of the embedding. The proof is in Appendix~\ref{app:proofs}; its main steps are showing $R(B,\cdot)$ is convex, $R'(B,0) > 0$, then using the convexity (and consequent non-decreasing derivative) to show that $R(B,\cdot)$ increases as $\lambda$ moves from 0 toward 1.

\subsection{Counterfactual hard-triplet generation and fine-tuning}
\label{sec:recipe}

Theorem~\ref{thm:hardtriplet} suggests that training on a distribution of hard triplets should implicitly cause the encoder to place less weight on the nuisance subspace. We now describe how we synthesize such a distribution and how we train on it.

To begin, we assemble a pool of 2{,}000 political and social issues from the Habermas Machine dataset~\citep{tessler2024ai} and Kialo,
\footnote{\url{https://huggingface.co/datasets/timchen0618/Kialo}} 
filtered by GPT-4o-mini for genuine policy debatability, and prompt Claude Sonnet 4 to generate 5 opinions per issue spanning the stance spectrum. We then sample anchors from this pool and prompt GPT-4o to rewrite each anchor into two versions. The \emph{preference match} preserves the anchor's stance and values but changes vocabulary, framing, and sentence structure. The \emph{semantic distractor} preserves much of the anchor's wording and structure but flips the stance. Together these form a triplet $(a,p,n)$ in which the preferred item is stance-aligned but less surface-similar, while the dispreferred item is surface-similar but stance-opposed. Applying the same rewrite prompt to real participant anchors from the 11 evaluation datasets produces the held-out hard triplets used in \S\ref{sec:hard} and \S\ref{sec:results}. All prompts and model details are in Appendix~\ref{app:prompts}.

We remark that the training data is completely synthetic and was generated independently of the 11 benchmark datasets used for evaluation. So, any performance gains on the natural data benchmark are a strong indicator of generalization. In addition, the way the triplets are generated is specifically designed to approximate the hard-triplet condition, $\E[\Delta_T\mid\mathcal{G}]\leq 0$.

We adapt the pretrained encoder $\psi$ with LoRA \citep{hu2022lora} ($r\!=\!16$, $\alpha\!=\!48$, targeting query and value projections). For a triplet $(a,p,n)$ we train with Bradley-Terry loss over cosine differences:
\[
\mathcal{L}_{\mathrm{BT}}(a,p,n) = \log\!\Big(1 + e^{-\big(\cos(\psi(a), \psi(p)) - \cos(\psi(a), \psi(n))\big)}\Big).
\]
Training and inference both use cosine geometry, so what we optimize is what is used downstream. We do not train to convergence on the hard-triplet distribution since this would lead to negative weight on the nuisance subspace instead of the desired invariance.

\section{Empirical Results}
\label{sec:results}

Fine-tuning substantially reduces the hard-triplet failure identified in \S\ref{sec:hard}. Our DPT-tuned sentence-T5-XL lifts hard-triplet (from the evaluation set) cosine accuracy from 48.3\% to 80.0\% and natural-triplet mean accuracy from 65.2\% to 68.6\%. Further, the recipe is not specific to sentence-T5: applying the same training to three widely-used BERT-derived encoders, e5-large-v2, BGE-large, and all-mpnet-base, produces similar patterns, with hard-triplet accuracy lifting by roughly 20 to 50 percentage points across models (Table~\ref{tab:cross-model}). The natural-data lift is significant at the participant level under a paired Wilcoxon test ($p = 3.4\times10^{-9}$, pooled over the $442$ test-split participants in the 11 datasets) and significant on $10$ of $11$ datasets under a triplet-level McNemar test (Appendix~\ref{app:paired-test}).

\begin{table}[ht]
\centering
\small
\caption{Base vs DPT-tuned cosine accuracy on the 11 natural-data datasets (test split) and on the 875 hard triplets across 5 training seeds. The recipe is hyperparameter-identical across models except for learning rate and hard-triplet count, which are val-selected per encoder.}
\label{tab:cross-model}
\begin{tabular*}{\textwidth}{@{\extracolsep{\fill}}lcccccc@{}}
\toprule
& \multicolumn{2}{c}{\textbf{Natural (\%)}} & \multicolumn{2}{c}{\textbf{Hard (\%)}} & \multicolumn{2}{c}{\textbf{Gain (pp)}} \\
\cmidrule(lr){2-3} \cmidrule(lr){4-5} \cmidrule(lr){6-7}
\textbf{Encoder} & base & tuned & base & tuned & nat. & hard \\
\midrule
sentence-T5-XL         & 65.2 & 68.6\,$\pm$\,0.2  &  48.3 & 80.0\,$\pm$\,0.4  & +3.4 & +31.7 \\
e5-large-v2            & 60.4 & 63.0\,$\pm$\,0.3  &  26.7 & 78.7\,$\pm$\,0.5  & +2.6 & +51.9 \\
BGE-large-en-v1.5      & 59.2 & 61.1\,$\pm$\,0.2  &   6.3 & 32.8\,$\pm$\,3.0  & +1.9 & +26.5 \\
all-mpnet-base-v2      & 57.8 & 58.6\,$\pm$\,0.2  &   8.2 & 27.4\,$\pm$\,2.2  & +0.8 & +19.2 \\
\bottomrule
\end{tabular*}
\end{table}

Table~\ref{tab:main} shows where the resulting DPT-tuned geometry sits relative to a broader set of off-the-shelf and stance-aware baselines, including methods explicitly designed for viewpoint or contradiction-sensitive similarity. It shows a representative cut of the full 25-model comparison (Appendix~\ref{app:full-table}). Our DPT-tuned sentence-T5-XL achieves the highest mean accuracy at 68.6\%, compared to 65.2\% for the next-best embedding (the untuned ST5-XL). Our method is best on 8 of 11 datasets, and no other model is best on more than one.

\begin{table}[ht]
\centering
\scriptsize
\caption{Triplet accuracy (\%) across the 11 evaluation datasets, abridged to seven representative models. A full comparison across 25 embedding models is in Appendix~\ref{app:full-table}.}
\label{tab:main}
\begin{tabular*}{\textwidth}{@{\extracolsep{\fill}}lccc|ccc|ccccc|c@{}}
\toprule
& \multicolumn{3}{c}{\textbf{GSC}} & \multicolumn{3}{c}{\textbf{Remesh}} & \multicolumn{5}{c}{\textbf{Polis}} & \\
\cmidrule(lr){2-4} \cmidrule(lr){5-7} \cmidrule(lr){8-12}
\textbf{Model} & AbG & AbV & Chat & Camp & Frgn & RTA & Seatl & BG & Brxt & Can & UBI & \textbf{Mean} \\
\midrule
ST5-XL + DPT (ours)            & \textbf{81.8} & \textbf{76.4} & 65.9 & \textbf{67.5} & \textbf{65.6} & 69.5 & \textbf{69.8} & \textbf{65.1} & \textbf{68.4} & \textbf{62.5} & 61.9 & \textbf{68.6} \\
ST5-XL                         & 81.1 & 70.5 & \textbf{69.7} & 65.8 & 61.1 & 69.9 & 60.7 & 57.6 & 56.4 & 55.5 & 69.2 & 65.2 \\
OAI text-emb-3-large           & 79.6 & 74.5 & 56.1 & 63.9 & 61.1 & \textbf{71.4} & 58.1 & 57.7 & 51.6 & 50.0 & 68.1 & 62.9 \\
ST5-Large                      & 75.5 & 65.5 & 57.2 & 64.6 & 59.9 & 69.2 & 58.8 & 58.1 & 49.9 & 56.8 & \textbf{72.2} & 62.5 \\
BGE-large-en                   & 64.7 & 59.5 & 56.5 & 62.5 & 57.6 & 68.0 & 54.2 & 55.5 & 53.2 & 52.9 & 66.3 & 59.2 \\
StanceAware-SBERT              & 58.5 & 58.2 & 36.4 & 64.8 & 59.7 & 68.2 & 68.1 & 56.1 & 52.5 & 52.3 & 68.3 & 58.5 \\
BGE-SparseCL (arguana)         & 62.4 & 61.7 & 42.7 & 59.2 & 57.2 & 66.1 & 50.1 & 56.1 & 57.5 & 52.6 & 63.0 & 57.2 \\
\bottomrule
\end{tabular*}
\end{table}

\section{Per-Topic Projected Embeddings}
\label{sec:probes}

Decorrelated preference tuning trains a single encoder whose cosine geometry is more preference-aware without using any data from the target topic. This section asks what changes when participant votes over statements are available, as is often the case for online deliberation platforms. The section serves three purposes. First, it gives a stronger topic-specific deployment mode where the base embedding model is held fixed and a low-rank linear map is learned from participant votes. Second, it tests the ideal-point model from \S\ref{sec:framework}. And third, it sharpens the diagnosis of cosine similarity on base embedding models. If a low-rank linear map works on a frozen encoder, it suggests that the encoder already contains a preference signal. DPT makes this structure more visible to cosine by implicitly down-weighting the nuisance component, and the per-topic map recovers it from votes.

\begin{table}[b]
\centering
\small
\caption{Each row relaxes one structural commitment of the ideal-point scorer; all are fit on frozen base sentence-T5-XL with Bradley-Terry loss across the 11 evaluation datasets (3-way participant split, val-selected hyperparameters, five seeds).}
\label{tab:probes}
\begin{tabular*}{\textwidth}{@{\extracolsep{\fill}}lccc@{}}
\toprule
\textbf{Scorer on frozen $\psi$} & \textbf{Form} & \textbf{Nat.\ (\%)} & \textbf{Hard (\%)} \\
\midrule
\textbf{Ideal-point scorer \emph{(ours)}}  & $-\lVert L^\top\psi(a) - L^\top\psi(x)\rVert^2$  & 77.6 $\pm$ 0.4 & 81.1 $\pm$ 0.5 \\
Nonlinear \textit{[drops linearity]}  & $-\lVert\phi(\psi(a)) - \phi(\psi(x))\rVert^2$ & 73.5 $\pm$ 0.6 & 66.5 $\pm$ 3.1 \\
Asymmetric \textit{[drops tied projection]}  & $-\lVert L_a^\top\psi(a) - L_x^\top\psi(x)\rVert^2$ & 77.0 $\pm$ 0.2 & 73.7 $\pm$ 0.4 \\
Inner product \textit{[drops distance]}  & $\langle L^\top\psi(a), L^\top\psi(x)\rangle$  & 77.9 $\pm$ 0.3 & 73.3 $\pm$ 0.2 \\

\bottomrule
\end{tabular*}
\end{table}

To begin, we fit the ideal-point utility directly on frozen sentence-T5-XL. Let $a$ be a participant $v$'s anchor and $x_j$ a candidate statement. We learn a rank-$r$ linear map $L^\top:\mathbb{R}^d\to\mathbb{R}^r$ and score candidates by squared Euclidean distance in the mapped space:
\small
\begin{align}
\label{eq:metric-scorer}
U(v,j)
= -\bigl\lVert L^\top\psi(a_v)-L^\top\psi(x_j)\bigr\rVert^2
= 2\langle L^\top\psi(a_v),L^\top\psi(x_j)\rangle
- \lVert L^\top\psi(x_j)\rVert^2
- \lVert L^\top\psi(a_v)\rVert^2.
\end{align}
\normalsize
We refer to $\tilde\psi(x) = L^\top\psi(x)$ as a \textit{projected embedding}. This form is the ideal-point model with the preference subspace and metric learned from votes. The same map is applied to anchors and candidates, and utility is distance rather than inner product. Fitting $L$ separately for each of the 11 deliberation datasets with Bradley-Terry loss on participant votes (3-way train/validation/test split, rank $r=20$, 5 seeds) reaches 77.6\% macro-mean accuracy over the datasets on the test split. Applied to the hard evaluation triplets, with no training on hard data, the same per-topic linear map reaches 81.1\% macro-mean accuracy (Table~\ref{tab:probes}, first row). This suggests that the frozen encoder already contains preference signal and that $L$ simply learns to read it out. This may also explain why DPT is so sample efficient --- the geometry only needs to be realigned rather than learned from scratch.

We now turn to the question of whether the model proposed in \S\ref{sec:framework} is empirically justified. The ideal-point scorer commits to three main structural claims about $\tilde\psi$: the scorer is \emph{linear} in $\psi$, \emph{tied} between anchor and item (both projected by the same $L$), and a \emph{distance} rather than an inner product. Table~\ref{tab:probes} relaxes each claim in turn. Replacing the linear $L^\top$ with a shared nonlinear MLP $\phi$ while keeping the squared-distance form costs 4.1\% on the natural data and 14.6\% on the hard evaluation triplets, indicating that $\psi$ is already rich enough that added per-text nonlinearity overfits rather than helps. Projecting anchor and item through independent $L_a, L_j$ matches natural-data accuracy but loses 7.4\% on the hard evaluation triplets. Dropping the item-quadratic $-\lVert L^\top\psi(j)\rVert^2$ for the bilinear cross-term alone (same rank, same shared $L$) leaves natural accuracy unchanged but loses 7.8\% on hard evaluation triplets. A possible explanation for this is that the item-norm encodes statement quality that directional similarity discards.

\begin{wrapfigure}{r}{0.42\textwidth}
\vspace{-1.0em}
\centering
\includegraphics[width=\linewidth]{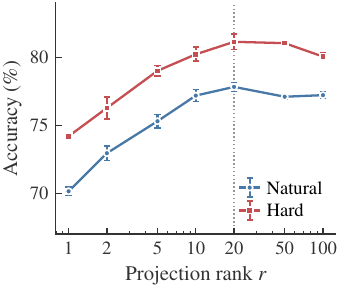}
\caption{Per-topic scorer accuracy versus projection rank $r$. Mean$\pm$std over five seeds, macro-averaged over 11 datasets.}
\label{fig:rank-saturation}
\vspace{-3.2em}
\end{wrapfigure}

Our model also assumes that the learned space is low-dimensional. To test this we swept across ranks $r\in\{1,2,5,10,20,50,100\}$. Figure~\ref{fig:rank-saturation} shows that accuracy rises from $r=1$ and plateaus at $r=20$. Per-topic supervision is data-efficient as well. Fixing $r=20$, the projected embedding crosses the universal DPT cosine at roughly 50 labeled triplets (Appendix~\ref{app:scorer-data-eff}).


\section{Discussion}
\label{sec:disc}
This paper develops methods for learning a preference geometry over text, motivated by the design choices that deliberative systems must make. A key aspect of deliberation is exposing participants to a range of viewpoints \citep{mutz2006hearing, fishkin2005experimenting,sunstein2002law}. Online platforms such as Polis, Remesh, Frankly, and the Stanford Online Deliberation Platform face this problem at scale, and must decide who should deliberate together, which comments to surface, and which viewpoints are missing \citep{small2021polis,fishkin2019deliberative,frankly2026}. A preference geometry gives one way to represent the viewpoints before making these decisions.

Two examples illustrate the point: forming deliberation groups, and aggregating the views expressed during deliberation. The composition of a deliberation group affects the quality of discussion, with quality highest at moderate levels of within-group disagreement \citep{esterling2015disagreement, karpowitz2007groups}. In practice, participants are assigned to small groups at random or by algorithms that balance demographics across groups alongside other objectives such as maximizing distinct pairwise meetings across rounds \citep{barrett2023submodular, barrett2024dream}, where demographic categories act as a coarse proxy for views \citep{mansbridge1999should}. \citet{yang2025bridging} move beyond demographics by using voting data as a proxy of views. A preference embedding space takes the final step, providing a direct measure of views in which objectives like balancing positions across rooms or controlling the level of within-room disagreement can be specified directly. Aggregation also has a natural formulation as voting in a metric space \citep{bulteau2021aggregation, feldman2016voting}. Since platforms typically surface several comments rather than one, the relevant problem is selecting a slate, which corresponds geometrically to clustering. Here $k$-median and $k$-center capture utilitarian and egalitarian aggregation, while proportional fairness requires any large, aligned subgroup to have a representative \citep{chen2019proportionally, micha2020proportionally, kellerhals2024proportional}.

A preference geometry can also aid generative social choice \citep{fish2026generative,boehmer2025gscnext}, where generative models expand the set of candidate statements. Before generation, it can identify groups who deserve representation, during generation it can guide the generative model, and after generation, it can measure how well a generated statement represents its intended group. Appendix~\ref{app:cluster-coherence} gives preliminary evidence that the tuned and projected geometries identify such groups better than the base embedding: on Remesh, users agree more with comments from their own cluster than from others, and the gap is significantly larger under the DPT-tuned and projected geometries.

\paragraph{Limitations and future work.} First, our evaluation focuses on within-participant rankings, not whether absolute distances are calibrated, and since utility is latent, direct verification is difficult. Yet, Bradley-Terry training does push correctly-ordered pairs apart by a margin that scales with preference strength, and Appendix~\ref{app:likert-correlation} shows that similarity in the tuned and projected spaces tracks continuous Likert ratings better than the base model.
Second, the gap between the universal tune and the per-topic projected embedding suggests that preferences have both a shared component across topics and a topic-specific component that only voting data on that topic recovers. A natural next step is to generate hard triplets conditioned on a target topic, producing a topic-specific embedding without per-topic votes.
Finally, an embedding aimed at capturing preferences is an empirical and imperfect representation of preferences. It can support sense-making and help surface diverse views, but it should not be treated as a perfect representation of any individual's considered judgment, and it should not be used as a basis for binding decisions \citep{revel2026ai}.

\section*{Acknowledgments}

This work was partially supported by the National Science Foundation under grant IIS-2229881; by the Office of Naval Research under grants N00014-24-1-2704 and N00014-25-1-2153; and by grants from the Cooperative AI Foundation and the Foresight Institute. This work was also partially supported by ONR MURI N00014-24-1-2742. Carter Blair is supported by an NSERC PGS D and a Cooperative AI PhD Fellowship. 

\section*{AI Acknowledgment}
The AI coding assistant Claude Code helped us write and debug code and run experiments, ChatGPT helped with the proof of Theorem~\ref{thm:hardtriplet}, and AI assistants helped with writing and editing. We take full responsibility for all content in this paper. LLMs that are part of the method are described in Appendix~\ref{app:prompts}.

\bibliographystyle{plainnat}
\bibliography{ref}

\appendix
\newpage

\section{Extended Related Work}
\label{app:related}

Work on collective decision-making over free-form text increasingly relies on some representation of participants or statements. Polis derives a low-dimensional opinion map from the participant-by-comment vote matrix via PCA and clustering \citep{small2021polis,small2023llmpolis}, generative social choice and PROSE group statements in an LLM-defined feature space to produce representative slates \citep{fish2026generative,boehmer2025gscnext}, \cite{blair2025approxconsensus} model approximate consensus as a region of sentence-embedding space, and \cite{de2026question} audit justified representation of question slates using cosine similarity of question embeddings as participant utility. Complementary work aggregates text without an explicit embedding geometry, for example through reward models, token-level policies, or sampling-based social-choice procedures \citep{tessler2024ai,blair2025generating,chooi2026finding,grandi2026socialchoicewithtext}.

A separate line of work studies the mismatch between semantic overlap and stance. Vahtola et al.\ \citep{vahtola-etal-2022-easy} show that generic sentence embeddings struggle when negation or antonymy flips meaning; Introne \citep{introne2023belief} and Ghafouri et al.\ \citep{ghafouri-etal-2024-love} fine-tune encoders to separate opposing viewpoints for stance detection or opinion retrieval; and counter-argument retrieval methods add explicit dissimilarity terms or sparsity-aware scoring to retrieve contradictions \citep{wachsmuth-etal-2018-retrieval,xu2024sparsecl}. We have a similar starting point, but make a unique diagnosis and propose a formal model, then use these to inspire a novel method.

A long tradition in political science estimates low-dimensional ideal points from political data. The canonical approach models legislators' roll-call votes as a function of their positions in a latent spatial model \citep{poole1985spatial,clinton2004statistical}. A parallel line recovers positions directly from text: Wordscores scales manifestos against reference documents \citep{laver2003extracting}, Wordfish estimates time-series party positions from speech \citep{slapin2008scaling}, and later work joins text with votes \citep{gerrish2011predicting}, combines topic models with ideal points \citep{vafa2020text}, or augments word embeddings with speaker metadata \citep{rheault2020word}. Our \S\ref{sec:probes} finding --- that preferences on a frozen sentence embedding are linearly accessible through a rank-$\sim$20 projection --- is consistent with this tradition, recovered here as a property of a general-purpose pretrained encoder on deliberation data from non-legislators.

Our work also connects to preference learning more broadly. Bradley-Terry objectives are standard in reward modeling and preference-based fine-tuning \citep{christiano2017deep,ouyang2022training,rafailov2023direct}, where they reshape a generative policy's output distribution. \citet{zhang2025beyondbt} also embed responses but score preferences with a skew-symmetric operator to express within-user cyclic preferences, a concern that does not arise in our cross-user setting. We use the Bradley-Terry objective to reshape a reusable embedding geometry that can be consumed by metric social choice and other downstream geometric procedures \citep{anshelevich2018metric,bulteau2021aggregation}, rather than only a per-query reward model.

\section{Dataset Details}
\label{app:datasets}

The GSC datasets come from two studies by \citet{fish2026generative}. In each, participants first write free-text opinions describing their views on a topic, then rate a set of AI-generated policy statements on a Likert scale. We construct triplets from all pairs of statements with distinct ratings for each participant. The abortion generation survey uses participants who authored the original opinions; the abortion validation survey uses a separate cohort who rate the same statements and give verbal feedback to the statements. The chatbot personalization survey follows the same structure. Data for the abortion surveys is available at \url{https://github.com/generative-social-choice/gsc_abortion} and for chatbot personalization at \url{https://github.com/generative-social-choice/chatbot_personalization}.

The Remesh datasets come from the Polarized Issues corpus, which collects binary agree/disagree votes on open-ended responses to political questions using the Remesh platform. Participants first write their own response to a prompt, then vote on responses written by other participants. We use three topics: campus protests, foreign intervention, and right to assemble. The data is available at \url{https://github.com/akonya/polarized-issues-data}.

The Polis datasets come from publicly available Polis conversations from \citet{small2021polis}. In Polis, participants write short comments and vote agree, disagree, or pass on comments written by others. We restrict to participants who both authored at least one comment and voted on at least five others, using their authored comments as anchors. These datasets have the shortest texts and sparsest preference signal of any platform in our evaluation. The data is available at \url{https://github.com/compdemocracy/openData}.

\paragraph{Licenses.} The GSC datasets are released under AGPL-3.0, the Polarized Issues (Remesh) data under CC-BY-4.0, and the Polis openData under CC-BY-4.0. The Habermas Machine dataset used for training-data generation is released under CC-BY-4.0 (data) and Apache-2.0 (code), and the Kialo dataset under MIT. The embedding models used are sentence-T5-XL (Apache-2.0), e5-large-v2 (MIT), BGE-large-en-v1.5 (MIT), and all-mpnet-base-v2 (Apache-2.0). All licenses permit research use.

\begin{table}[t]
\centering
\small
\caption{Evaluation datasets. Participants authored anchors (own text) and voted on statements (others' text); pairwise preference triplets derive from their vote orderings. GSC uses a small fixed statement pool; Remesh and Polis use open pools seeded by participants themselves. We use all 11 datasets for evaluation.}
\label{tab:datasets}
\begin{tabular*}{\textwidth}{@{\extracolsep{\fill}}lllrrrr@{}}
\toprule
Source & Dataset & Abbreviation & Participants & Anchors & Statements & Triplets \\
\midrule
GSC & Abortion (generation users) & AbG & 100 & 500 & 5 & 846 \\
 & Abortion (validation users) & AbV & 100 & 993 & 10 & 3,513 \\
 & Chatbot Personalization & Chat & 100 & 788 & 6 & 1,104 \\
Remesh & Campus Protests & Camp & 298 & 1,176 & 1,011 & 21,989 \\
 & Foreign Intervention & Frgn & 289 & 857 & 764 & 12,428 \\
 & Right to Assemble & RTA & 295 & 1,445 & 1,274 & 28,340 \\
Polis & Seattle \$15/hour & Seatl & 13 & 13 & 29 & 1,103 \\
 & Bowling Green AA & BG & 222 & 531 & 607 & 1,363,191 \\
 & Brexit Consensus & Brxt & 11 & 50 & 50 & 2,468 \\
 & Canadian Reform & Can & 19 & 150 & 151 & 19,366 \\
 & NZ UBI & UBI & 15 & 15 & 51 & 4,643 \\
\midrule
\multicolumn{2}{l}{Total} & & 1,462 & 6,458 & 3,958 & 1,458,991 \\
\bottomrule
\end{tabular*}
\end{table}

\section{Proof of Theorem~\ref{thm:hardtriplet}}
\label{app:proofs}

Fix a symmetric in-subspace operator $B$ with $B=P_SBP_S$, and write
\[
\ell(u)=\log(1+e^{-u}).
\]
For a triplet $(a,p,n)$, recall that
\[
\Delta_B
=
\psi_S(a)^\top B\bigl(\psi_S(p)-\psi_S(n)\bigr),
  \qquad
  \Delta_T
  =
  \bigl\langle \psi_\perp(a),\psi_\perp(p)-\psi_\perp(n)\bigr\rangle.
\]
Thus, the risk can be written as
\[
  R(B,\lambda)
  =
  \mathbb{E}\bigl[\ell(\Delta_B+\lambda\Delta_T)\bigr].
\]

Our goal is to show that $R(B,\lambda)<R(B,1)$ for every $\lambda\in[0,1)$.

To do so, we will first note bounds which will be required below. Since the encoder outputs unit-norm embeddings and orthogonal projection cannot increase the norm (by the Pythagorean theorem),
\[
  \|\psi_\perp(a)\|\le 1,
  \qquad
  \|\psi_\perp(p)-\psi_\perp(n)\|\le
  \|\psi_\perp(p)\|+\|\psi_\perp(n)\|\le 2.
\]
Therefore,
\[
  |\Delta_T|
  =
  \bigl|
  \langle \psi_\perp(a),\psi_\perp(p)-\psi_\perp(n)\rangle
  \bigr|
  \le 2
\]
everywhere. Also, since $B$ is fixed and the embeddings have norm at most one,
\[
  |\Delta_B|
  \le 2\|B\|_{\mathrm{op}}.
\]
Since $|\Delta_B+\lambda\Delta_T|\le 2\|B\|_{\mathrm{op}}+2|\lambda|$ for each fixed $\lambda$, and $\ell$ is continuous, $\ell(\Delta_B+\lambda\Delta_T)$ is bounded and therefore integrable.

We now justify differentiating $R(B,\lambda)$ with respect to $\lambda$. The derivative of the logistic loss is
\[
  \ell'(u)
  =
  -\frac{1}{1+e^u},
\]

so $|\ell'(u)|\le 1$ for every $u\in\mathbb{R}$. Fix $\lambda\in\mathbb{R}$ and, for nonzero $h$, define the difference quotient random variable
\[
  Q_h
  =
  \frac{
    \ell(\Delta_B+(\lambda+h)\Delta_T)
    -
    \ell(\Delta_B+\lambda\Delta_T)
  }{h}.
\]
For each realization of the triplet, $\Delta_B$ and $\Delta_T$ are fixed finite numbers, and the map
\[
  \eta\mapsto \ell(\Delta_B+\eta\Delta_T)
\]
is differentiable. Hence, as $h\to 0$,
\[
  Q_h
  \to
  \ell'(\Delta_B+\lambda\Delta_T)\Delta_T
\]
pointwise. Moreover, by the mean value theorem, for each $h\neq 0$ there exists some $\xi_h$ between $\lambda$ and $\lambda+h$ such that
\[
  Q_h
  =
  \ell'(\Delta_B+\xi_h\Delta_T)\Delta_T.
\]
Therefore
\[
  |Q_h|
  \le
  |\ell'(\Delta_B+\xi_h\Delta_T)|\,|\Delta_T|
  \le 2,
\]
using $|\ell'|\le 1$ and $|\Delta_T|\le 2$. The constant function $2$ is integrable, so dominated convergence gives
\[
  R'(B,\lambda)
  =
  \lim_{h\to 0}
  \frac{R(B,\lambda+h)-R(B,\lambda)}{h}
  =
  \lim_{h\to 0}\mathbb{E}[Q_h]
  =
  \mathbb{E}\bigl[
    \ell'(\Delta_B+\lambda\Delta_T)\Delta_T
  \bigr].
\]

Thus, for every $\lambda\in\mathbb{R}$,
\[
  R'(B,\lambda)
  =
  \mathbb{E}\bigl[
    \ell'(\Delta_B+\lambda\Delta_T)\Delta_T
  \bigr].
\]

We now evaluate this derivative at $\lambda=0$. Let
\[
  \mathcal{G}
  =
  \sigma\bigl(\psi_S(a),\psi_S(p),\psi_S(n)\bigr)
\]
be the $\sigma$-algebra generated by the preference-subspace parts of the triplet. Since $B=P_SBP_S$, the quantity $\Delta_B$ is a function only of the projected variables in $\mathcal{G}$. Hence $\ell'(\Delta_B)$ is $\mathcal{G}$-measurable. By the tower property,

\[
  R'(B,0)
  =
  \mathbb{E}\bigl[\ell'(\Delta_B)\Delta_T\bigr]
  =
  \mathbb{E}\Bigl[
    \mathbb{E}\bigl[
      \ell'(\Delta_B)\Delta_T
      \mid \mathcal{G}
    \bigr]
  \Bigr].
\]

Since $\ell'(\Delta_B)$ is $\mathcal{G}$-measurable, the pull-out property gives
\[
  \mathbb{E}\bigl[
    \ell'(\Delta_B)\Delta_T
    \mid \mathcal{G}
  \bigr]
  =
  \ell'(\Delta_B)\,
  \mathbb{E}[\Delta_T\mid\mathcal{G}].
\]

Therefore
\[
  R'(B,0)
  =
  \mathbb{E}\Bigl[
    \ell'(\Delta_B)\,
    \mathbb{E}[\Delta_T\mid\mathcal{G}]
  \Bigr].
\]

The hard-triplet condition states that
\[
  \mathbb{E}[\Delta_T\mid\mathcal{G}]\le 0
  \quad\text{a.s.},
\]
with strict inequality on a set of positive probability. Also,
\[
  \ell'(u)
  =
  -\frac{1}{1+e^u}
  <0
  \qquad
  \text{for every }u\in\mathbb{R},
\]
and hence $\ell'(\Delta_B)<0$ everywhere. Therefore the product
\[
  \ell'(\Delta_B)\,
  \mathbb{E}[\Delta_T\mid\mathcal{G}]
\]
is nonnegative almost surely and strictly positive on a set of positive probability. Hence
\[
  R'(B,0)>0.
\]

It remains to show that this positivity at zero implies the desired comparison with $\lambda=1$. The loss $\ell$ is convex because
\[
  \ell''(u)
  =
  \frac{e^u}{(1+e^u)^2}
  >0
  \qquad
  \text{for every }u\in\mathbb{R}.
\]
For each realization of the triplet, the map
\[
  \lambda\mapsto \Delta_B+\lambda\Delta_T
\]
is affine, so
\[
  \lambda\mapsto \ell(\Delta_B+\lambda\Delta_T)
\]
is convex. Taking expectations preserves convexity, and hence $R(B,\cdot)$ is convex. Since $R(B,\cdot)$ is differentiable, its derivative is nondecreasing. Therefore, for every $\lambda\in[0,1]$,
\[
  R'(B,\lambda)\ge R'(B,0)>0.
\]

Now fix any $\bar\lambda\in[0,1)$. By the mean value theorem, there exists some $c\in(\bar\lambda,1)$ such that
\[
  R(B,1)-R(B,\bar\lambda)
  =
  R'(B,c)(1-\bar\lambda).
\]
Since $c\in[0,1]$, we have $R'(B,c)>0$, and since $1-\bar\lambda>0$,
\[
  R(B,1)-R(B,\bar\lambda)>0.
\]
Equivalently,
\[
  R(B,\bar\lambda)<R(B,1).
\]
Because $\bar\lambda\in[0,1)$ was arbitrary,
\[
  R(B,\lambda)<R(B,1)
  \qquad
  \text{for every }\lambda\in[0,1).
\]
\qed

\section{Additional Experimental Details}
\label{app:additional}

\paragraph{Hyperparameter selection.} The sentence-T5-XL recipe reported in the main text was selected by a sweep over learning rate $\in \{1.25\text{e-}5, 5\text{e-}5, 1.25\text{e-}4, 5\text{e-}4\}$, hard-triplet count $\in \{250, 500, 750, 1000, 2000\}$, LoRA rank $\in \{4, 8, 16, 32, 64\}$, and LoRA $\alpha = 3r$ on a held-out validation split. The selected configuration is $\mathrm{lr} = 1.25\text{e-}4$, $n_{\mathrm{hard}} = 750$, $r = 16$, $\alpha = 48$, which corresponds to 46 gradient steps at batch size 16 on a random 750 of the 10{,}000-triplet synthetic pool, sampled uniformly without replacement and resampled independently for each training seed. For cross-model transfer (e5, BGE, all-mpnet) we fix rank and alpha and re-select $(\mathrm{lr}, n_{\mathrm{hard}})$ per encoder from the same grid; selected configurations are $(5\text{e-}5, 1000)$ for e5-large, $(1.25\text{e-}4, 500)$ for BGE-large, and $(1.25\text{e-}4, 1000)$ for all-mpnet. We AdamW with a linear-decay schedule and a 10\% warmup ratio for all models.

\paragraph{Compute.} All experiments were run on a single NVIDIA A100 GPU. Training a single DPT seed takes less than 20 minutes.

\section{Additional Experiments}

\subsection{Error Decomposition}
\label{app:errors}

To understand the residual errors of the tuned model, we classify a stratified sample of 913 errors using GPT-4o. Each error is a triplet where the tuned model incorrectly ranks the dispreferred statement higher. We present the full triplet (anchor texts, preferred, dispreferred) and ask the classifier to select from six categories, presented in randomized order to avoid position bias.

\begin{table*}[t]
\centering
\small
\caption{LLM-classified error categories for tuned model errors (913 triplets, stratified across datasets, capped at 5 per participant).}
\label{tab:errors}
\begin{tabular*}{\textwidth}{@{\extracolsep{\fill}}lrr@{}}
\toprule
\textbf{Category} & \textbf{Count} & \textbf{\%} \\
\midrule
Surface similarity & 557 & 61.0 \\
Insufficient anchor signal & 187 & 20.5 \\
Subtle value distinction & 154 & 16.9 \\
Both options plausible & 10 & 1.1 \\
Style/register mismatch & 3 & 0.3 \\
None of the above & 2 & 0.2 \\
\bottomrule
\end{tabular*}
\end{table*}

Surface similarity remains the dominant failure mode at 61\%, which indicates room for further improvement through refined hard-triplet training. Insufficient anchor signal accounts for 20.5\% and is concentrated in Polis datasets where comments are short: 73\% of Bowling Green errors fall in this category, compared to 6\% for GSC. Subtle value distinctions at 16.9\% represent cases where both options broadly match the anchor's views but differ on fine-grained nuance below the resolution of sentence-level similarity. The high number of ``insufficient anchor signal'' errors is notable since this is technically something no embedding model can solve. This must be addressed upstream by encouraging participants to write more or perhaps by giving them scaffolding to aid in their articulation of their preferences \citep{blair2025reflective, li2025eliciting, handa2024bayesian}.

\subsection{LoRA Rank Ablation.} We vary LoRA rank $r \in \{4, 8, 16, 32, 64\}$ with $\alpha = 3r$, holding all other hyperparameters at the selected ST5-XL configuration. Table~\ref{tab:rank-ablation} reports held-out test accuracy averaged across the 11 main evaluation datasets. Accuracy peaks at $r{=}16$: $67.6\%$ at $r{=}4$, $68.7\%$ at $r{=}16$, $67.1\%$ at $r{=}32$, and $63.8\%$ at $r{=}64$. With only 46 optimizer steps on 750 triplets, higher ranks introduce more free parameters than the data can constrain, and the extra capacity starts overwriting the pretrained geometry that the tuning is meant to realign, not replace.

\begin{table}[h]
\centering
\small
\caption{LoRA rank ablation on sentence-T5-XL (single seed, mean test accuracy across 11 datasets).}
\label{tab:rank-ablation}
\begin{tabular*}{\textwidth}{@{\extracolsep{\fill}}lccccc@{}}
\toprule
LoRA rank $r$ & 4 & 8 & 16 & 32 & 64 \\
\midrule
Mean test acc.\ (\%) & 67.6 & 68.3 & \textbf{68.7} & 67.1 & 63.8 \\
\bottomrule
\end{tabular*}
\end{table}

\subsection{Loss Ablation: Bradley-Terry vs InfoNCE.}
We replace the pairwise Bradley-Terry loss with InfoNCE (temperature $0.05$). For each anchor in a batch of 16, the positive is its own preference match and the negatives are its own semantic distractor plus the other 15 anchors' preference matches as in-batch negatives, giving 16 negatives per anchor. All other hyperparameters are held at the selected ST5-XL configuration. The two losses yield essentially identical mean test accuracy (68.7\% for BT vs 68.6\% for InfoNCE), so the gain comes from the counterfactual triplet construction itself, not the choice of pairwise objective. We report Bradley-Terry in the main results because its score is exactly the signed cosine difference used at inference, and because it operates on each triplet independently — no batch size or batch composition to tune. 

\subsection{Normal-Correlation Triplet Ablation.}
\label{app:normal-triplet}
The hard-triplet construction in \S\ref{sec:hard} engineers a worst-case correlation between stance and wording so that the nuisance margin satisfies the hard-triplet condition $\E[\Delta_T \mid \mathcal{G}] \leq 0$. The purpose of this is to force the model to rely less on the nuisance subspace and more on the preference subspace. We test whether this engineering is necessary by generating $2{,}000$ \emph{normal-correlation} training triplets in which stance and wording move together. This mimics more traditional NLI pair training where the correlation between the desired signal and a nuisance signal is not explicitly broken \citep{gao2021simcse}. The rewrite prompt mirrors the hard prompt but flips the role assignment, so the preference match shares vocabulary with the anchor while the semantic distractor uses different vocabulary. The full prompt is in Appendix~\ref{app:prompts}. By construction $\E[\Delta_T \mid \mathcal{G}] > 0$, the opposite of the hard-triplet condition used in Theorem~\ref{thm:hardtriplet}. We sweep the same hyperparameter grid used for the hard-triplet selection in Appendix~\ref{app:additional} and select the configuration with the highest validation macro-mean accuracy on the 11 natural datasets, which gives $\mathrm{lr} = 1.25\text{e-}5$, $n = 2000$, $r = 64$. Re-running this configuration over five seeds yields 67.0 $\pm$ 0.1\% mean test accuracy and 59.3 $\pm$ 0.2\% accuracy on the hard evaluation triplets. The matched hard-trained DPT configuration in Table~\ref{tab:cross-model} reaches 68.6 $\pm$ 0.2\% test and 80.0 $\pm$ 0.4\% on the hard evaluation triplets. Normal-correlation training therefore recovers a small lift over base cosine, but it is significantly worse than DPT.

\subsection{Hard-Positive vs.\ Hard-Negative Ablation.}
\label{app:polarity}
Each hard triplet has two hard sides: the preference match keeps the anchor's stance in new wording (a hard positive), and the semantic distractor keeps the anchor's wording but reverses its stance (a hard negative). To measure what each side contributes, we fix a random sample of 2{,}000 anchors from the training pool. For each anchor we make one additional GPT-4o call with a mirrored prompt (Appendix~\ref{app:prompts}), which produces an \emph{easy positive} (same stance, nearly the same wording) and an \emph{easy negative} (opposite stance, different wording). We then recombine the four rewrites so that one side changes at a time: \emph{hard negative only} (anchor, easy positive, hard negative), where both candidates are lexically close to the anchor; \emph{hard positive only} (anchor, hard positive, easy negative), where both candidates are lexically distant; \emph{both sides hard}, our standard construction; and \emph{easy--easy} (anchor, easy positive, easy negative). Mean word-level Jaccard overlap with the anchor confirms the manipulation: it is 0.12 for hard positives versus 0.91 for easy positives, and 0.70 for hard negatives versus 0.12 for easy negatives. Every condition uses the selected sentence-T5-XL configuration (750 triplets sampled per seed from the 2{,}000, 46 steps) and is evaluated exactly as in the main text. We use five seeds.

\begin{table}[h]
\centering
\small
\caption{Hard-positive vs.\ hard-negative ablation on sentence-T5-XL (same 2{,}000 anchors in every row, five seeds). Natural is the mean test accuracy over the 11 datasets; Hard is accuracy on the held-out hard evaluation triplets.}
\label{tab:polarity}
\begin{tabular*}{\textwidth}{@{\extracolsep{\fill}}lcccc@{}}
\toprule
\textbf{Training triplets} & \textbf{Positive} & \textbf{Negative} & \textbf{Nat.\ (\%)} & \textbf{Hard (\%)} \\
\midrule
Both sides hard \emph{(ours)} & hard & hard & \textbf{68.5} $\pm$ 0.1 & \textbf{79.9} $\pm$ 0.6 \\
Hard negative only & easy & hard & 66.8 $\pm$ 0.2 & 78.9 $\pm$ 0.4 \\
Hard positive only & hard & easy & 66.0 $\pm$ 0.2 & 74.1 $\pm$ 0.5 \\
Easy--easy & easy & easy & 64.3 $\pm$ 0.1 & 71.1 $\pm$ 0.8 \\
\midrule
Untuned sentence-T5-XL & -- & -- & 65.2 & 48.3 \\
\bottomrule
\end{tabular*}
\end{table}

Table~\ref{tab:polarity} shows that both sides hard is best on both metrics. The hard negative carries most of the hard-triplet gain: hard-negative-only training reaches 78.9\% hard accuracy, against 79.9\% for both sides and 74.1\% for hard-positive-only. On natural data, however, neither side alone matches the full construction, which is 1.7 points above hard-negative-only and 2.5 points above hard-positive-only. Each hard side also improves on easy--easy training on both metrics, so the hard positive contributes even though the hard negative contributes more. Easy--easy triplets resemble the normal-correlation triplets of Appendix~\ref{app:normal-triplet}, but here they are trained at the DPT configuration rather than at a configuration selected for them, so the two sets of numbers differ.

\subsection{Sensitivity to the Rewrite Prompt and Generator.}
\label{app:generator}
The hard-triplet rewrite asks the generator to balance two goals at once (preserve or reverse the stance, while controlling word overlap), so we test whether DPT depends on the exact wording of the prompt in Appendix~\ref{app:prompts} or on GPT-4o. Using the same 2{,}000 anchors as Appendix~\ref{app:polarity}, we regenerate both rewrites under four configurations: GPT-4o with two alternative prompts, and the original prompt with GPT-4o-mini and with Claude Sonnet 4.6. Prompt B restates the task in prose, presents the preference match first, uses a different worked example, and frames the two rewrites as statements the person would agree with or reject. Prompt C is a terse benchmark-builder instruction with no worked example and no language about values. Both prompts are given in Appendix~\ref{app:prompts}. The control is the original GPT-4o rewrites of the same 2{,}000 anchors, which is also the both-sides-hard row of Table~\ref{tab:polarity}. Training and evaluation are as in Appendix~\ref{app:polarity}.

\begin{table}[h]
\centering
\small
\caption{Sensitivity of DPT to the rewrite prompt and generator (sentence-T5-XL, same 2{,}000 anchors in every row, five seeds). Overlap is the mean word-level Jaccard overlap of the preference match and of the semantic distractor with the anchor.}
\label{tab:generator}
\begin{tabular*}{\textwidth}{@{\extracolsep{\fill}}llccc@{}}
\toprule
\textbf{Generator} & \textbf{Prompt} & \textbf{Overlap (match / distr.)} & \textbf{Nat.\ (\%)} & \textbf{Hard (\%)} \\
\midrule
GPT-4o & original (control) & 0.12 / 0.70 & 68.5 $\pm$ 0.1 & 79.9 $\pm$ 0.6 \\
GPT-4o & B & 0.15 / 0.70 & 68.3 $\pm$ 0.1 & 80.6 $\pm$ 0.2 \\
GPT-4o & C & 0.19 / 0.54 & 67.6 $\pm$ 0.1 & 79.2 $\pm$ 0.4 \\
GPT-4o-mini & original & 0.16 / 0.55 & 67.9 $\pm$ 0.2 & 79.9 $\pm$ 0.2 \\
Claude Sonnet 4.6 & original & 0.07 / 0.71 & 68.1 $\pm$ 0.2 & 80.3 $\pm$ 0.4 \\
\midrule
Untuned sentence-T5-XL & -- & -- & 65.2 & 48.3 \\
\bottomrule
\end{tabular*}
\end{table}

Across all five configurations, hard-triplet accuracy stays between 79.2\% and 80.6\%, within about one seed standard deviation of the control (79.9 $\pm$ 0.6\%). Natural accuracy is at most 0.9 points below the control (prompt C), and every configuration remains well above the untuned encoder. The two configurations whose distractors share the least wording with the anchor (prompt C and GPT-4o-mini) have the lowest natural accuracy, consistent with the construction working through the wording contrast. Both the much smaller GPT-4o-mini and a model from a different family reproduce the control's hard-triplet accuracy, so DPT depends neither on the exact prompt nor on a single generator's notion of an opposite stance.

\subsection{Robustness to Label Noise.}
\label{app:label-noise}
LLM-generated triplets will sometimes be wrong, for example when the generator fails to reverse the stance of the semantic distractor. We test robustness to such errors directly. Starting from the selected sentence-T5-XL configuration, we swap the preference match and the semantic distractor in a random $\rho \in \{10, 30, 50, 70, 90\}\%$ of the 750 training triplets, chosen independently for each seed, and leave everything else unchanged. We use five seeds per level.

\begin{table}[h]
\centering
\small
\caption{DPT under label noise (sentence-T5-XL, five seeds). A fraction $\rho$ of the training triplets has its preference match and semantic distractor swapped. The $\rho = 0$ row is the DPT model of Table~\ref{tab:cross-model}.}
\label{tab:label-noise}
\begin{tabular*}{\textwidth}{@{\extracolsep{\fill}}lcc@{}}
\toprule
\textbf{Label noise $\rho$} & \textbf{Nat.\ (\%)} & \textbf{Hard (\%)} \\
\midrule
0\% & 68.6 $\pm$ 0.2 & 80.0 $\pm$ 0.4 \\
10\% & 68.5 $\pm$ 0.3 & 77.1 $\pm$ 0.8 \\
30\% & 67.8 $\pm$ 0.4 & 64.1 $\pm$ 1.9 \\
50\% & 65.6 $\pm$ 0.4 & 48.6 $\pm$ 1.1 \\
70\% & 63.3 $\pm$ 0.4 & 34.6 $\pm$ 2.3 \\
90\% & 61.2 $\pm$ 0.3 & 17.3 $\pm$ 1.4 \\
\midrule
Untuned sentence-T5-XL & 65.2 & 48.3 \\
\bottomrule
\end{tabular*}
\end{table}

Performance degrades smoothly and monotonically as noise increases (Table~\ref{tab:label-noise}). At 10\% noise, natural accuracy is unchanged (68.5\% vs.\ 68.6\%) and hard-triplet accuracy falls by 2.9 points. At exactly 50\%, where the labels carry no information about which candidate is preferred, DPT returns to the untuned encoder (65.6\% vs.\ 65.2\% natural, 48.6\% vs.\ 48.3\% hard). It falls below the untuned encoder only once most labels are inverted. This is consistent with Theorem~\ref{thm:hardtriplet}, which requires the hard-triplet condition only in conditional expectation rather than on every triplet, so a minority of flawed triplets weakens the effect without reversing it.

\subsection{Participant-Level Significance.}
\label{app:paired-test}
Table~\ref{tab:cross-model} reports dataset-mean accuracies across five training seeds, which is the right unit for comparing tuning recipes but does not directly speak to whether the gain is reliable on the unit of deployment, the individual participant. We therefore run a paired test on per-participant accuracies. For each of the 11 evaluation datasets we replay the test-split scoring loop under both base sentence-T5-XL and the DPT-tuned encoder, keeping the per-triplet outcome ($1$ if the encoder ranks the preference match above the distractor, $0$ otherwise) and the per-participant accuracy (e.g., the accuracy for participant $i$ using the base model is $a^{\text{base}}_i$). Each participant then has a pair $(a^{\text{base}}_i, a^{\text{tuned}}_i)$, and each triplet has a pair of binary outcomes. We report a paired Wilcoxon signed-rank test on the participant accuracies and a McNemar exact test on the triplet outcomes.

Pooled across all 442 participants in the 11 test cohorts, the mean accuracy lift is +2.0 pp (median +0.9 pp). Tuned beats base on 51.6\% of participants, ties on 17.2\%, and loses on 31.2\%. The paired Wilcoxon test yields $p = 3.4 \times 10^{-9}$ and the paired t-test yields $p = 1.8 \times 10^{-8}$. McNemar over the 24,834 discordant triplets returns $b = 14{,}891$ wins for tuned and $c = 9{,}943$ wins for base, with $p = 9.5 \times 10^{-218}$. Per-dataset results are in Table~\ref{tab:paired-test}. The participant-level Wilcoxon reaches $p < 0.05$ on $5$ of $11$ datasets, while the triplet-level McNemar reaches $p < 0.05$ on $10$ of $11$. The lone non-mover under both is GSC chatbot, where DPT shows the smallest accuracy lift across the entire benchmark ($\Delta = +0.4$ pp). The Wilcoxon non-significance on the four small Polis cohorts (Seattle, Brexit, Canadian, UBI) reflects power rather than effect, since each shows tuned-mean above base-mean and the triplet-level McNemar reaches $p < 0.05$ on all four. The per-dataset and pooled accuracies in Table~\ref{tab:paired-test} are means of per-participant accuracies, which weight each participant equally rather than each triplet (as Table~\ref{tab:full-main} and Table~\ref{tab:main} do).

\begin{table}[h]
\centering
\small
\caption{Paired test of DPT-tuned vs base sentence-T5-XL at the participant and triplet level (test split, seed-$42$ DPT). $\Delta$ is the per-participant accuracy difference in percentage points. Wilcoxon is the paired signed-rank test on participant accuracies, McNemar is the exact binomial test on per-triplet wins and losses.}
\label{tab:paired-test}
\begin{tabular*}{\textwidth}{@{\extracolsep{\fill}}lcccccc@{}}
\toprule
\textbf{Dataset} & $n$ & base (\%) & tuned (\%) & $\Delta$ (pp) & Wilcoxon $p$ & McNemar $p$ \\
\midrule
GSC Abortion (gen)        &  30 & 78.1 & 84.5 & $+6.4$ & $4.3{\times}10^{-2}$ & $1.5{\times}10^{-3}$ \\
GSC Abortion (val)        &  30 & 71.8 & 78.0 & $+6.2$ & $9.9{\times}10^{-3}$ & $3.5{\times}10^{-8}$ \\
GSC Chatbot (gen)         &  30 & 65.1 & 65.5 & $+0.4$ & $7.5{\times}10^{-1}$ & $1.0$ \\
Remesh Campus Protests    &  90 & 63.7 & 65.9 & $+2.2$ & $3.5{\times}10^{-5}$ & $1.3{\times}10^{-12}$ \\
Remesh Foreign Interv.    &  87 & 62.2 & 63.3 & $+1.1$ & $4.3{\times}10^{-2}$ & $3.4{\times}10^{-3}$ \\
Remesh Right to Assemble  &  89 & 69.3 & 70.3 & $+1.0$ & $7.5{\times}10^{-2}$ & $1.4{\times}10^{-2}$ \\
Polis Seattle             &   4 & 70.4 & 72.9 & $+2.6$ & $7.5{\times}10^{-1}$ & $4.1{\times}10^{-2}$ \\
Polis Bowling Green       &  67 & 61.1 & 61.8 & $+0.7$ & $3.0{\times}10^{-2}$ & $7.2{\times}10^{-189}$ \\
Polis Brexit              &   4 & 69.9 & 72.0 & $+2.1$ & $0.62$ & $5.4{\times}10^{-3}$ \\
Polis Canadian            &   6 & 50.7 & 57.1 & $+6.3$ & $6.2{\times}10^{-2}$ & $6.9{\times}10^{-10}$ \\
Polis UBI                 &   5 & 55.0 & 58.0 & $+3.0$ & $0.12$ & $1.4{\times}10^{-2}$ \\
\midrule
\textbf{Pooled}           & \textbf{442} & \textbf{65.6} & \textbf{67.6} & $\boldsymbol{+2.0}$ & $\boldsymbol{3.4{\times}10^{-9}}$ & $\boldsymbol{9.5{\times}10^{-218}}$ \\
\bottomrule
\end{tabular*}
\end{table}

\subsection{Per-Topic Scorer: Rank Sweep}
\label{app:scorer-rank}

Figure~\ref{fig:rank-saturation} in \S\ref{sec:probes} shows accuracy as a function of projection rank $r$. Table~\ref{tab:rank-saturation} reports the exact macro-mean values with standard deviations across five seeds.

\begin{table}[h]
\centering
\small
\caption{Per-topic scorer accuracy as a function of rank (macro-mean over 11 datasets, mean$\pm$std over five seeds).}
\label{tab:rank-saturation}
\begin{tabular*}{\textwidth}{@{\extracolsep{\fill}}lccccccc@{}}
\toprule
Rank $r$     & 1 & 2 & 5 & 10 & 20 & 50 & 100 \\
\midrule
Natural (\%) & 70.1 $\pm$ 0.3 & 72.9 $\pm$ 0.6 & 75.4 $\pm$ 0.5 & 77.1 $\pm$ 0.4 & \textbf{77.6} $\pm$ 0.4 & 77.1 $\pm$ 0.2 & 77.3 $\pm$ 0.2 \\
Hard (\%)    & 74.0 $\pm$ 0.4 & 76.4 $\pm$ 0.8 & 79.2 $\pm$ 0.4 & 80.4 $\pm$ 0.6 & \textbf{81.1} $\pm$ 0.5 & 81.0 $\pm$ 0.5 & 80.3 $\pm$ 0.4 \\

\bottomrule
\end{tabular*}
\end{table}

\subsection{Per-Topic Scorer: Data Efficiency}
\label{app:scorer-data-eff}

\begin{figure}[h]
\centering
\includegraphics[width=\linewidth]{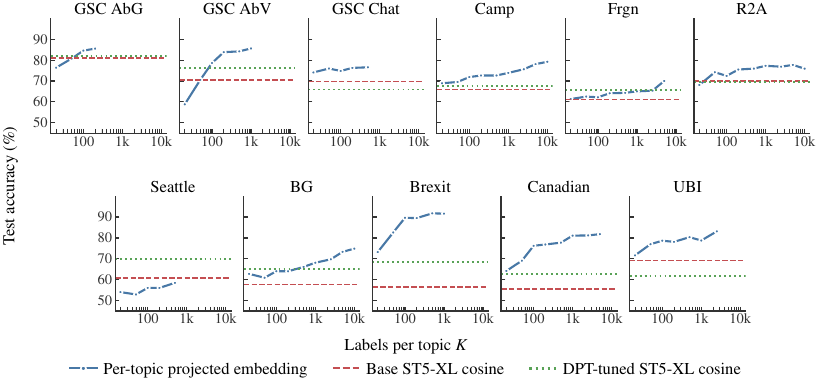}
\caption{Data efficiency of the rank-20 per-topic projected embedding on base sentence-T5-XL.}
\label{fig:data-eff}
\end{figure}

Fixing $r{=}20$, we vary the number of labeled triplets per topic and evaluate held-out natural accuracy. Figure~\ref{fig:data-eff} plots the learning curve across individual datasets and the macro-mean against the base and universal DPT-tuned cosine baselines. Averaging over datasets, $K{\approx}50$ labeled triplets already cross the universal DPT baseline, and accuracy saturates by $K{\approx}1{,}000$.

\subsection{User Clustering Coherence on Remesh}
\label{app:cluster-coherence}

Here we test whether the DPT-tuned and per-topic projected embeddings provide a geometry in which clusters are coherent in approval behavior, beyond what the base ST5-XL geometry already provides. The Remesh deliberation transcripts are well suited to this question as each participant authors one or more comments, votes \emph{Agree} or \emph{Disagree} on a sample of comments written by others, and the comments tend to be more substantive than those from Polis. Further, in contrast to GSC, there are many statements that were voted on so the approval level for any cluster can be estimated. Taken together, this allows us to get another angle on how well the geometry aligns with participants' expressed approval preferences in a setting akin to how it may be deployed. 

For each participant $u$ we form a user vector by
mean-pooling the encoder embeddings of their authored comments. We do this under three geometries, the base sentence-T5-XL encoder, the DPT-tuned encoder of \S\ref{sec:method}, and the rank-20 per-topic projection $L^\top \psi$ of \S\ref{sec:probes}, with $L$ taken from the ideal-point scorer fit on the same dataset. We then run $k$-means on the user vectors for $k\in\{3,5,8,10\}$. For every user $u$ assigned to cluster $c$, we compute their \emph{within-cluster approval rate}, defined as the fraction
of \emph{Agree} votes among the cluster-$c$-authored
comments that $u$ actually voted on. The denominator counts only votes that exist, and comments that were never shown to $u$ are excluded. The analogous \emph{across-cluster} rate uses the comments not in $u$'s cluster.

We aggregate per-user rates within a cluster by
vote-weighted mean and then macro-average over the clusters at a given $k$. The headline statistic is the lift $\Delta = \overline{\text{within}} - \overline{\text{across}}$, which is approximately zero under random assignment. Each cell in Table~\ref{tab:cluster-coherence}, indexed by
dataset, encoder, and $k$, averages over five $k$-means seeds. We additionally compute a permutation null by shuffling cluster labels, with 50 permutations per seed. All shuffle lifts have absolute value at most 0.005, so we report raw lifts directly.

\begin{table}[h]
\centering
\small
\caption{Within$-$across approval-rate lift $\Delta$ (\%) on Remesh user clusters, by encoder (base ST5-XL; DPT-tuned ST5-XL; rank-20 per-topic projection of base ST5-XL via $L^\top\psi$). Each cell is the mean of five $k$-means seeds; per-seed shuffle null has $|\Delta|\le 0.5\%$. Best entry per row in \textbf{bold}.}
\label{tab:cluster-coherence}
\begin{tabular*}{\textwidth}{@{\extracolsep{\fill}}llccc@{}}
\toprule
\textbf{Dataset} & $k$ & \textbf{Base} & \textbf{Tuned} & \textbf{Proj} \\
\midrule
\multirow{4}{*}{Campus Protests}
 &  3 & 9.7 & \textbf{11.1} & 9.5 \\
 &  5 & 6.4 & 8.5          & \textbf{11.5} \\
 &  8 & 8.7 & 11.0          & \textbf{11.1} \\
 & 10 & 8.8 & 10.4          & \textbf{10.7} \\
\midrule
\multirow{4}{*}{Foreign Intervention}
 &  3 & 2.4 & 5.8 & \textbf{10.1} \\
 &  5 & 6.4 & 5.1 & \textbf{11.4} \\
 &  8 & 7.7 & \textbf{9.0} & 0.7 \\
 & 10 & 6.0 & \textbf{7.1} & 2.6 \\
\midrule
\multirow{4}{*}{Right to Assemble}
 &  3 & 1.4 & 1.8          & \textbf{2.0} \\
 &  5 & 2.8 & \textbf{3.7} & 2.9 \\
 &  8 & \textbf{4.5} & 3.3 & 2.3 \\
 & 10 & 2.1 & 3.4          & \textbf{3.9} \\
\bottomrule
\end{tabular*}
\end{table}

Table~\ref{tab:cluster-coherence} reports the
per-configuration lifts. Both adapted geometries exceed the base on a strong majority of the 12 cells. Clustering separates users most cleanly on Campus Protests, where $\Delta \approx 0.10$ for both the tuned encoder and the projection at all values of $k$. It separates them less
cleanly on Foreign Intervention, where the projection collapses at $k\ge 8$, and only weakly on Right to Assemble, where the global within-rate is approximately 0.78 and there is little disagreement left to partition.

Aggregating across the 12 configurations, the mean lift is 5.6\% for base, 6.7\% for DPT-tuned, and 6.6\% for the per-topic projected embedding. The mean rank, with lower being better, is 2.50, 1.75, and 1.75 respectively. Paired comparisons over configurations give
$\Delta_{\text{DPT-tuned}}-\Delta_{\text{base}} = +1.11$ pp, with 95\% bootstrap CI $[+0.4,+1.8]$ and Wilcoxon $p=0.016$, and $\Delta_{\text{projected}}-\Delta_{\text{base}} = +1.00$ pp, with CI $[-1.3, +3.1]$ and $p=0.38$. The wider interval for the projection is driven by the two values of $k$ on
Foreign Intervention where the rank-20 projection
collapses. DPT-tuned and the projected embedding are statistically indistinguishable at this scale, with $\Delta = +0.1$ pp, CI $[-1.9, +2.3]$, and $p=0.97$. These results suggest that the DPT-tuned embedding provides a geometry which would be useful for the types of tasks identified in \S\ref{sec:disc}.

\subsection{Likert-Rating Correlation on GSC}
\label{app:likert-correlation}

The triplet metric used throughout the paper collapses each participant's continuous Likert rating of a candidate statement into a binary preference between two statements for the GSC datasets.
However, the GSC surveys release the underlying ratings on a 0 to 6 scale, which lets us ask the more direct question of how strongly cosine similarity between a participant's free-text opinion and a candidate statement tracks the rating they actually gave it, which can give us some indication of whether distances in the DPT-tuned and per-topic projected embedding track latent utility as we would hope.

For each of the three GSC surveys, we recover from the raw CSVs every $(u, s, r)$ triple in which participant $u$ rated statement $s$ with Likert score $r \in [0, 6]$. We assemble an anchor pool per participant from all of their free-text responses, which include the open-ended opinion text in the generation and chatbot surveys together with the per-rating justification text that all three surveys collect. The validation cohort writes only justifications. When correlating with the rating of statement $s$, we exclude the user's justification text written about $s$ from the anchor pool. The remaining texts are mean-pooled to give a single anchor embedding per rating row. We then compute Spearman rank correlation between cosine similarity and Likert rating, pooled across all rating rows in a survey, under three geometries. These are the base sentence-T5-XL encoder, the DPT-tuned encoder of \S\ref{sec:method}, and the rank-20 per-topic projected embedding $L^\top \psi$ of
\S\ref{sec:probes}, with $L$ fit from the ideal-point
scorer on the same survey's votes.

Table~\ref{tab:likert-correlation} shows that both adapted geometries lift the pooled correlation substantially over the base encoder on every survey. On the two abortion surveys, which are the most explicitly preference-laden topic in GSC, the per-topic projected embedding more than doubles the chatbot-survey correlation gap and pushes pooled correlation past 0.7 on the validation cohort. The chatbot survey gives the smallest absolute correlations across the board, and it is the one survey on which the global tune beats the per-topic projection.

\begin{table}[h]
\centering
\small
\caption{Pooled Spearman rank correlation between cosine similarity (anchor $\to$ statement) and the participant's Likert rating, across the three GSC surveys. Higher is better. Best per row in \textbf{bold}.}
\label{tab:likert-correlation}
\begin{tabular*}{\textwidth}{@{\extracolsep{\fill}}lcccc@{}}
\toprule
\textbf{Survey} & \textbf{$n$ ratings} & \textbf{Base} & \textbf{Tuned} & \textbf{Proj} \\
\midrule
GSC Abortion (gen) &  500 & 0.436 & 0.614          & \textbf{0.626} \\
GSC Abortion (val) & 1000 & 0.415 & 0.544          & \textbf{0.721} \\
GSC Chatbot  & 1100 & 0.254 & \textbf{0.385} & 0.331 \\
\bottomrule
\end{tabular*}
\end{table}

If we accept that Likert ratings give us a noisy view of the participants' latent utility, then this experiment provides some evidence that the distances in the DPT-tuned embedding and the per-topic projected embedding are more closely related to utility than distances in the base embedding model.

\subsection{Stacking the Per-Topic Probe on the Tuned Encoder}
\label{app:probe-on-tuned}

The per-topic metric scorer of \S\ref{sec:probes} is fit on the base sentence-T5-XL encoder. A natural follow-up question is whether the rank-20 projection would perform better if trained on the encoder that has been globally tuned by DPT. We re-run the same val-selected metric probe pipeline on the DPT-tuned encoder, holding all other choices fixed, including the rank, the validation grid, the three-way participant split, and the five seeds.

The natural-test macro is essentially unchanged, moving from 77.6 (base ST5-XL) to 78.0 (DPT-tuned). The hard evaluation triplet performance lifts from 81.1 (base ST5-XL) to 87.4 (DPT-tuned), a 6.3 percentage-point gain, with non-negative gains on every dataset and seven datasets gaining over five points.

\begin{table}[h]
\centering
\small
\caption{Per-dataset macro-mean test and 875-triplet hard accuracy of the rank-20 metric probe, fit at val-selected hyperparameters on either the base encoder or the DPT-tuned encoder. Mean and standard deviation over five seeds. Best per row in \textbf{bold}.}
\label{tab:probe-on-tuned}
\begin{tabular*}{\textwidth}{@{\extracolsep{\fill}}lcccc@{}}
\toprule
\textbf{Dataset} & \textbf{Test (base)} & \textbf{Test (tuned)} & \textbf{Hard (base)} & \textbf{Hard (tuned)} \\
\midrule
GSC Abortion (gen) & \textbf{88.3 $\pm$ 0.8} & 87.5 $\pm$ 0.3 & 78.0 $\pm$ 0.0 & \textbf{79.2 $\pm$ 0.4} \\
GSC Abortion (val) & 85.5 $\pm$ 2.1 & \textbf{86.3 $\pm$ 0.5} & 82.8 $\pm$ 2.7 & \textbf{86.8 $\pm$ 0.4} \\
GSC Chatbot       & \textbf{76.7 $\pm$ 1.5} & 76.0 $\pm$ 1.2 & 67.5 $\pm$ 1.3 & \textbf{77.4 $\pm$ 0.6} \\
Polis Seattle     & 74.6 $\pm$ 4.0 & \textbf{76.2 $\pm$ 0.8} & 98.5 $\pm$ 3.4 & \textbf{100.0 $\pm$ 0.0} \\
Polis Bowling Green & 72.2 $\pm$ 0.3 & \textbf{73.9 $\pm$ 0.8} & 84.2 $\pm$ 1.3 & \textbf{93.8 $\pm$ 2.2} \\
Polis Brexit      & 83.4 $\pm$ 1.7 & \textbf{91.5 $\pm$ 1.0} & 79.2 $\pm$ 1.5 & \textbf{88.8 $\pm$ 1.9} \\
Polis Canadian    & \textbf{75.7 $\pm$ 1.1} & 75.5 $\pm$ 1.0 & 77.4 $\pm$ 1.5 & \textbf{85.2 $\pm$ 1.8} \\
Polis UBI         & \textbf{78.4 $\pm$ 0.7} & 72.1 $\pm$ 0.3 & 86.7 $\pm$ 0.0 & 86.7 $\pm$ 0.0 \\
Remesh Campus     & \textbf{76.9 $\pm$ 0.2} & 76.5 $\pm$ 0.3 & 85.4 $\pm$ 1.1 & \textbf{92.2 $\pm$ 0.4} \\
Remesh Foreign    & \textbf{67.0 $\pm$ 0.1} & 66.5 $\pm$ 0.2 & 80.8 $\pm$ 1.8 & \textbf{88.0 $\pm$ 0.7} \\
Remesh RTA        & 74.9 $\pm$ 0.5 & \textbf{76.4 $\pm$ 0.1} & 71.8 $\pm$ 0.8 & \textbf{83.6 $\pm$ 0.5} \\
\midrule
\textbf{Macro mean} & 77.6 & \textbf{78.0} & 81.1 & \textbf{87.4} \\
\bottomrule
\end{tabular*}
\end{table}

This asymmetry between natural and hard data can be explained using our formal framework from \S\ref{sec:framework}. The ideal-point scorer projects into a learned rank-20 subspace and never sees the orthogonal complement, so the cosine decomposition of \eqref{eq:decomp} does not literally apply to the probe's score, but the same mechanism reappears inside the projection. Our hypothesis is that val-selection on natural data tends to pull a few directions into the column space of $L$ that are not strictly preference-aligned, because on natural data those directions correlate positively with preference and so improve validation accuracy. They are nuisance directions that look like signal when semantic and preferential similarity are correlated. When the resulting probe is then evaluated on hard triplets, those nuisance directions contribute against preference, and the projected embedding pays for them with degraded hard-triplet accuracy.

By suppressing the out-of-subspace nuisance globally, the DPT-tuning leaves val-selection with cleaner candidate directions to choose among, and the resulting tuned $L$ contains less of the regime-dependent contamination. We see this directly in the geometry of the learned subspaces. Letting $Q_{\text{base}}$ and $Q_{\text{tuned}}$ denote orthonormal bases of the two probes' columns, the singular values of $Q_{\text{base}}^\top Q_{\text{tuned}}$ are the cosines of the principal angles between the two rank-20 subspaces. Across the eleven datasets, the largest cosine averages 0.95 and the median averages 0.80, while the smallest averages 0.17. The two probes share a core preference subspace where most directions are well aligned, and each has a small number of directions that are nearly orthogonal to anything in the other. Those near-orthogonal residues are the encoder-specific component of each fit, and on hard evaluation triplets it is likely the case that the tuned encoder's residue is benign while the base encoder's residue carries the nuisance signal that costs accuracy.

A second fingerprint comes from decomposing the score margin as $2 \langle L^\top \psi(a),\, L^\top(\psi(p) - \psi(n)) \rangle + \lVert L^\top \psi(n) \rVert^2 - \lVert L^\top \psi(p) \rVert^2$, an in-projected-space inner product margin plus a projected item-norm difference, reported per dataset in Table~\ref{tab:probe-on-tuned-margin}. The projected embedding on top of the DPT-tuned model improves hard accuracy on 10 of 11 datasets, and on 8 of those the tuned inner product margin is $1.8$ to $5.9$ times larger than the base, with the item-norm term staying small in magnitude relative to the inner product margin. The remaining two, Polis Bowling Green and Remesh Foreign, also gain hard accuracy but with only a marginal increase in the mean inner product margin (ratios $1.1$ and $1.3$ respectively), indicating the improvement there is in where the margin lands rather than its mean. The takeaway is that the same rank-20 budget extracts a stronger preference signal once the base directions have been cleaned by DPT.

\begin{table}[h]
\centering
\small
\caption{Per-dataset mean of the in-projected-space inner product margin $2 \langle L^\top \psi(a), L^\top(\psi(p) - \psi(n)) \rangle$ and the projected item-norm term $\lVert L^\top \psi(n) \rVert^2 - \lVert L^\top \psi(p) \rVert^2$ on the dataset's share of the 875 hard triplets, computed under the rank-20 metric probe fit on either the base or the tuned encoder (single seed). The ratio column is tuned divided by base on the inner product margin.}
\label{tab:probe-on-tuned-margin}
\begin{tabular*}{\textwidth}{@{\extracolsep{\fill}}lccccc@{}}
\toprule
& \multicolumn{3}{c}{\textbf{Inner product margin}} & \multicolumn{2}{c}{\textbf{Item-norm difference}} \\
\cmidrule(lr){2-4} \cmidrule(lr){5-6}
\textbf{Dataset} & \textbf{Base} & \textbf{Tuned} & \textbf{Ratio} & \textbf{Base} & \textbf{Tuned} \\
\midrule
GSC Abortion (gen)  & +1.30 & +2.38 & 1.8 & $-0.25$ & $-0.39$ \\
GSC Abortion (val)  & +1.96 & +3.47 & 1.8 & $-0.54$ & $-0.61$ \\
GSC Chatbot         & +0.11 & +0.64 & 5.9 & $+0.02$ & $+0.10$ \\
Polis Seattle       & +0.93 & +1.94 & 2.1 & $+0.08$ & $+0.17$ \\
Polis Bowling Green & +1.17 & +1.34 & 1.1 & $-0.40$ & $-0.09$ \\
Polis Brexit        & +0.56 & +2.17 & 3.9 & $-0.02$ & $-0.42$ \\
Polis Canadian      & +1.16 & +5.11 & 4.4 & $-0.02$ & $-0.93$ \\
Polis UBI           & +1.11 & +2.00 & 1.8 & $-0.18$ & $-0.08$ \\
Remesh Campus       & +1.79 & +3.28 & 1.8 & $-0.27$ & $-0.08$ \\
Remesh Foreign      & +2.17 & +2.71 & 1.3 & $-0.71$ & $-0.50$ \\
Remesh RTA          & +0.82 & +1.66 & 2.0 & $-0.15$ & $+0.07$ \\
\bottomrule
\end{tabular*}
\end{table}

Two practical readings follow. First, DPT and the per-topic probe are complementary contributions, not redundant. DPT shifts the global encoder geometry to suppress the nuisance subspace, the per-topic projected embedding extracts the dataset-specific preference subspace, and the two compose with no measurable cost on natural data and a consistent gain on hard triplets. Second, val-selection on natural data is not an unalloyed good for downstream evaluation under distribution shift. When the candidate directions a fit can choose from include some that correlate with preference, val-selection can pull them in, and the model will look fine on its own held-out split while degrading on any distribution where the nuisance correlation flips.

\section{Full Model Comparison}
\label{app:full-table}

Table~\ref{tab:full-main} is the complete version of the abridged main-results table (Table~\ref{tab:main} in \S\ref{sec:results}), covering all 25 embedding models evaluated in this work.

The 25 baselines are: ST5-Base, ST5-Large, ST5-XL \citep{ni2022sentence}; all-mpnet-base-v2, all-MiniLM-L6, all-MiniLM-L12, all-distilroberta, NLI-mpnet, Paraphrase-MiniLM \citep{reimers2019sentence}; e5-large-v2 \citep{wang2022e5}; BGE-large-en-v1.5 \citep{xiao2024cpack}; GTE-large \citep{li2023gte}; mxbai-embed-large-v1 \citep{lee2024mxbai}; Arctic-embed-l \citep{merrick2024arctic}; Stella-en-1.5B \citep{zhang2024stella}; Qwen2-1.5B-instruct \citep{yang2024qwen2}; OpenAI text-embedding-3-small and text-embedding-3-large \citep{openai2024embeddings}; voyage-3 \citep{voyage2024voyage3}; voyage-3-large \citep{voyage2025voyage3large}; voyage-4, voyage-4-lite, voyage-4-large \citep{voyage2026voyage4}; StanceAware-SBERT \citep{ghafouri-etal-2024-love}; and BGE-SparseCL (arguana) \citep{xu2024sparsecl}.

\begin{table*}[h]
\centering
\scriptsize
\caption{Triplet accuracy (\%) across all 11 evaluation datasets, full comparison across 25 models. Best result in each column in \textbf{bold}. Abridged version in the body as Table~\ref{tab:main}.}
\label{tab:full-main}
\begin{tabular*}{\textwidth}{@{\extracolsep{\fill}}lccc|ccc|ccccc|c@{}}
\toprule
& \multicolumn{3}{c}{\textbf{GSC}} & \multicolumn{3}{c}{\textbf{Remesh}} & \multicolumn{5}{c}{\textbf{Polis}} & \\
\cmidrule(lr){2-4} \cmidrule(lr){5-7} \cmidrule(lr){8-12}
\textbf{Model} & AbG & AbV & Chat & Camp & Frgn & RTA & Seatl & BG & Brxt & Can & UBI & \textbf{Mean} \\
\midrule
ST5-XL + DPT (ours)            & \textbf{81.8} & \textbf{76.4} & 65.9 & \textbf{67.5} & \textbf{65.6} & 69.5 & \textbf{69.8} & \textbf{65.1} & \textbf{68.4} & \textbf{62.5} & 61.9 & \textbf{68.6} \\
ST5-XL                         & 81.1 & 70.5 & \textbf{69.7} & 65.8 & 61.1 & 69.9 & 60.7 & 57.6 & 56.4 & 55.5 & 69.2 & 65.2 \\
OAI text-emb-3-large           & 79.6 & 74.5 & 56.1 & 63.9 & 61.1 & \textbf{71.4} & 58.1 & 57.7 & 51.6 & 50.0 & 68.1 & 62.9 \\
ST5-Large                      & 75.5 & 65.5 & 57.2 & 64.6 & 59.9 & 69.2 & 58.8 & 58.1 & 49.9 & 56.8 & \textbf{72.2} & 62.5 \\
Voyage-4-large                 & 69.9 & 69.0 & 63.7 & 60.0 & 57.9 & 69.0 & 55.4 & 58.3 & 50.3 & 48.3 & 63.6 & 60.5 \\
e5-large-v2                    & 63.9 & 63.6 & 51.8 & 65.1 & 59.4 & 70.5 & 53.5 & 57.1 & 56.2 & 57.5 & 66.1 & 60.4 \\
Voyage-4                       & 72.3 & 66.6 & 62.1 & 58.4 & 56.7 & 67.0 & 54.9 & 57.4 & 50.8 & 50.1 & 64.5 & 60.1 \\
ST5-Base                       & 68.6 & 62.5 & 47.1 & 63.7 & 59.0 & 66.6 & 58.4 & 56.9 & 51.1 & 56.0 & 68.8 & 59.9 \\
NLI-mpnet                      & 56.1 & 66.9 & 47.6 & 65.6 & 61.0 & 69.5 & 58.3 & 56.9 & 52.8 & 55.5 & 68.7 & 59.9 \\
MxBAI-embed-large              & 63.9 & 62.4 & 55.1 & 62.7 & 59.9 & 69.2 & 54.2 & 56.2 & 50.5 & 54.5 & 70.0 & 59.9 \\
GTE-large                      & 67.6 & 65.6 & 59.5 & 60.2 & 56.3 & 67.7 & 54.7 & 55.5 & 46.2 & 51.6 & 69.5 & 59.5 \\
OAI text-emb-3-small           & 63.8 & 65.3 & 51.9 & 62.1 & 58.2 & 70.8 & 56.2 & 57.0 & 50.3 & 50.4 & 66.0 & 59.3 \\
Voyage-3-large                 & 68.3 & 61.0 & 60.6 & 57.2 & 57.5 & 67.6 & 58.5 & 57.6 & 50.7 & 49.8 & 63.5 & 59.3 \\
Voyage-4-lite                  & 70.1 & 64.5 & 61.3 & 58.2 & 56.8 & 65.4 & 53.3 & 57.0 & 51.2 & 49.3 & 65.5 & 59.3 \\
BGE-large-en                   & 64.7 & 59.5 & 56.5 & 62.5 & 57.6 & 68.0 & 54.2 & 55.5 & 53.2 & 52.9 & 66.3 & 59.2 \\
StanceAware-SBERT              & 58.5 & 58.2 & 36.4 & 64.8 & 59.7 & 68.2 & 68.1 & 56.1 & 52.5 & 52.3 & 68.3 & 58.5 \\
Paraphrase-MiniLM              & 58.2 & 67.8 & 51.7 & 59.7 & 58.3 & 67.5 & 51.7 & 56.4 & 53.3 & 52.4 & 64.7 & 58.3 \\
all-mpnet-base                 & 57.1 & 55.6 & 60.4 & 58.5 & 57.0 & 69.2 & 54.4 & 56.4 & 47.1 & 52.4 & 68.0 & 57.8 \\
Arctic-embed-l                 & 56.4 & 62.4 & 58.5 & 57.6 & 56.2 & 67.5 & 53.4 & 57.6 & 50.2 & 47.3 & 64.2 & 57.4 \\
all-distilroberta              & 61.8 & 55.8 & 53.4 & 59.5 & 57.5 & 67.6 & 53.7 & 55.4 & 50.5 & 48.5 & 65.8 & 57.2 \\
BGE-SparseCL (arguana)         & 62.4 & 61.7 & 42.7 & 59.2 & 57.2 & 66.1 & 50.1 & 56.1 & 57.5 & 52.6 & 63.0 & 57.2 \\
Voyage-3                       & 63.8 & 50.4 & 57.5 & 58.0 & 54.9 & 65.2 & 53.0 & 55.5 & 50.6 & 45.8 & 65.2 & 56.4 \\
all-MiniLM-L12                 & 57.8 & 57.1 & 47.6 & 59.2 & 57.8 & 67.2 & 52.0 & 55.3 & 49.9 & 48.8 & 64.1 & 56.1 \\
all-MiniLM-L6                  & 55.4 & 52.3 & 48.9 & 60.2 & 57.6 & 67.1 & 53.6 & 55.1 & 48.4 & 48.1 & 65.4 & 55.7 \\
Qwen2-1.5B-instruct            & 59.6 & 50.8 & 49.7 & 56.5 & 54.9 & 58.0 & 48.5 & 50.9 & 49.4 & 50.5 & 62.7 & 53.8 \\
Stella-en-1.5B                 & 43.5 & 50.2 & 65.3 & 55.0 & 55.1 & 54.0 & 53.4 & 49.1 & 47.4 & 57.0 & 49.7 & 52.7 \\
\bottomrule
\end{tabular*}
\end{table*}

\subsection{Baseline encoding format}
\label{app:baseline-format}

Many baselines were trained as asymmetric retrieval encoders, expecting the query side and the passage side of an input to be formatted differently. Treating these encoders symmetrically (passing both anchor and candidate through plain \texttt{encode(text)}) consistently under-represents what they can do. We therefore apply each model's intended query/passage convention, treating the participant's anchor text as the query and the candidate statement as the passage. The conventions below are sourced from each model card or registered \texttt{config\_sentence\_transformers.json}, and in all cases cosine similarity is computed between the query-side and passage-side embeddings.

\begin{itemize}[noitemsep, topsep=0pt, leftmargin=*]
  \item \textbf{Symmetric (no prefixes):} ST5-XL/Large/Base, all-mpnet-base, all-MiniLM-\{L6,L12\}, all-distilroberta, NLI-mpnet, Paraphrase-MiniLM, GTE-large \citep{li2023gte} (the older non-instruct variant), OpenAI text-embedding-3-\{small, large\}, StanceAware-SBERT.
  \item \textbf{e5-large-v2:} \texttt{"query: "} prefix on anchors, \texttt{"passage: "} prefix on candidates.
  \item \textbf{BGE-large-en-v1.5, MxBAI-embed-large-v1:} \texttt{"Represent this sentence for searching relevant passages: "} prefix on anchors, plain on candidates.
  \item \textbf{Snowflake Arctic-embed-l-v2.0:} \texttt{"query: "} prefix on anchors, plain on candidates. The v2 redesign uses E5-style prefixes rather than the v1 BGE prompt.
  \item \textbf{BGE-SparseCL (arguana):} plain encoding on both sides, matching the authors' own evaluation script. The released checkpoint ships without an SBERT \texttt{modules.json}; we verified that the default sentence-transformers wrapper applies mean pooling, matching the paper's training-time \texttt{-{}-pooler\_type avg}.
  \item \textbf{Voyage:} the API exposes an \texttt{input\_type} parameter; we pass \texttt{"query"} for anchors and \texttt{"document"} for candidates.
  \item \textbf{Instruction-tuned (GTE-Qwen2-1.5B-instruct, Stella-en-1.5B-v5):} task-tailored instruction wrapped in each model's standard \texttt{"Instruct: <task>$\backslash$nQuery: "} template on anchors, plain text on candidates.
\end{itemize}

\paragraph{Instruction-tuned prompt selection.}
The registered prompts in each model's \texttt{config\_sentence\_transformers.json} target web-search retrieval (e.g., \emph{``Given a web search query, retrieve relevant passages that answer the query''}), which does not match a sentence-to-sentence preference task. Qwen was trained with diverse task instructions and supports arbitrary task strings. Stella was trained primarily on two registered prompts, \texttt{s2p\_query} for sentence-to-passage retrieval and \texttt{s2s\_query} for sentence-to-sentence similarity, with the model card recommending those two for general use. We swap in a task-tailored instruction, \emph{``Given an opinion statement on a contested social or political issue, retrieve other statements that share the same stance and underlying values,''} using the template described above.

For Stella, we verified that this choice is benign on natural data: the task-tailored instruction yields a 52.7\% macro-mean, the registered \texttt{s2s\_query} prompt yields 52.8\% (a 0.1 pp difference), and the \texttt{s2p\_query} prompt yields 49.4\% (worse, as expected for the web-search framing). For Qwen, the task-tailored instruction and the registered web-search instruction give identical 53.8\% macro-mean. The Table~\ref{tab:full-main} rows for these two models report the task-tailored configuration.

\section{Prompts}
\label{app:prompts}

Three LLM prompts define the training-data pipeline: GPT-4o-mini filters the issue pool, Claude Sonnet generates 5 diverse opinions per surviving issue, and GPT-4o rewrites a sampled anchor into a hard triplet. The same rewrite prompt is used to produce the held-out hard eval triplets from real participant anchors (\S\ref{sec:hard}). The robustness experiments in Appendices~\ref{app:polarity} and~\ref{app:generator} use three further rewrite prompts, given at the end of this section. Placeholders in braces (\texttt{\{issue\_text\}}, \texttt{\{anchor\}}) are substituted at call time. For reproducibility, exact model identifiers and sampling settings are:

\begin{itemize}[leftmargin=2em,itemsep=1pt,topsep=2pt]
\item \textbf{Issue filter:} \texttt{gpt-4o-mini}, temperature $0$, max tokens $5$.
\item \textbf{Opinion generation:} \texttt{claude-sonnet-4-20250514}, temperature $1.0$, max tokens $1024$.
\item \textbf{Hard-triplet rewrite:} \texttt{gpt-4o}, temperature $0.7$, max tokens $400$.
\item \textbf{Rewrite variants (Appendices~\ref{app:polarity} and~\ref{app:generator}):} \texttt{gpt-4o} (prompts B and C, mirrored prompt), \texttt{gpt-4o-mini} and \texttt{claude-sonnet-4-6} (original rewrite prompt); temperature $0.7$, max tokens $400$.
\end{itemize}

\paragraph{Issue filter (GPT-4o-mini, one call per candidate issue).}
{\footnotesize
\begin{verbatim}
You are deciding whether a debate topic is a political, policy, or
social issue suitable for studying public opinion and preference
diversity.

ACCEPT if:
- It is a political, policy, economic, or social issue
- People's positions reflect their values, ideology, or worldview
  (not just personal taste)
- It is the kind of topic debated in legislatures, newspapers, public
  surveys, or town halls
- There is a clear spectrum of positions (e.g., pro-regulation vs.
  free-market)

REJECT if:
- It is purely factual or scientific with a clear answer
- It is abstract philosophy with no policy relevance
- It is a niche technical question that only specialists would engage
  with
- It is about personal lifestyle preferences rather than public affairs
- It has near-universal consensus
- It is nonsensical or too vague

Topic: "{issue_text}"

Respond with exactly one word: ACCEPT or REJECT
\end{verbatim}
}

\paragraph{Opinion generation (Claude Sonnet, one call per issue).}
{\footnotesize
\begin{verbatim}
Topic: {issue_text}

Write 5 standalone opinions on this topic from 5 different people.
These should read like things people wrote in a forum - NOT like
they are answering a question.

The 5 opinions should be clearly distinct and evenly spaced across
the full opinion spectrum:
1. Strongly supportive - fully committed to this position
2. Moderately supportive - in favor but with reservations
3. Genuinely ambivalent - sees valid points on both sides
4. Moderately opposed - against but acknowledges some merit
5. Strongly opposed - firmly against

Important:
- Do not start with Yes, No, I agree, I disagree or any
  response-like framing
- Each opinion should stand alone as a statement of belief
- Make the difference between adjacent positions (e.g., 1 vs 2,
  4 vs 5) clear and meaningful
- Keep them short and natural - average ~25 words, some shorter,
  some longer

{"opinions": ["op1", "op2", "op3", "op4", "op5"]}
\end{verbatim}
}

\paragraph{Hard-triplet rewrite (GPT-4o, one call per sampled anchor).}
{\footnotesize
\begin{verbatim}
You are given a person's opinion statement on a political or social
topic.

Original statement: "{anchor}"

Generate two rewritten versions that test whether a model understands
values vs. word overlap:

1. SEMANTIC DISTRACTOR (tricks models that rely on word overlap):
   - Keep nearly IDENTICAL wording and sentence structure as the
     original
   - Only flip the conclusion/stance to be the opposite
   - Goal: MAXIMIZE word overlap while having opposite meaning
   - Example: "Renewable energy must be prioritized above all else"
     -> "Renewable energy is worth considering, but should NOT be
     prioritized above all else"

2. PREFERENCE MATCH (requires understanding values, not words):
   - Express the SAME stance and values as the original
   - But use completely DIFFERENT vocabulary, framing, and sentence
     structure
   - Goal: MINIMIZE word overlap while preserving the underlying
     position
   - Example: "Renewable energy must be prioritized above all else"
     -> "Fossil fuel dependence is our greatest threat - government
     must act decisively on clean power"

Respond with valid JSON only:
{"semantic_distractor": "...", "preference_match": "..."}
\end{verbatim}
}

\paragraph{Normal-correlation rewrite (GPT-4o, one call per sampled anchor).}
{\footnotesize
\begin{verbatim}
You are given a person's opinion statement on a political or social
topic.

Original statement: "{anchor}"

Generate two rewritten versions that exhibit the natural correlation
between stance and wording, where same stance tends to share words
and opposite stance tends to use different words:

1. PREFERENCE MATCH (high word overlap with the anchor, same stance):
   - Reuse most of the anchor's vocabulary and content words
   - Preserve the original stance and values
   - BUT it must be a genuine paraphrase, NOT a verbatim or near-
     verbatim copy of the anchor. Change at least 3-5 words to
     synonyms, rearrange clauses, or restructure the sentence.
     Otherwise the rewrite is invalid.
   - Goal: MAXIMIZE word overlap while still meaningfully rewording
   - Example: "Renewable energy must be prioritized above all else"
     -> "We must put renewable energy ahead of every other concern"

2. SEMANTIC DISTRACTOR (low word overlap with the anchor, opposite
   stance):
   - Use completely DIFFERENT vocabulary, framing, and sentence
     structure
   - Express the OPPOSITE stance on the same underlying issue
   - Goal: MINIMIZE word overlap while having opposite meaning
   - Example: "Renewable energy must be prioritized above all else"
     -> "Cheap, reliable fossil fuels are what people actually
     need. Chasing wind and solar comes at too great a cost."

Respond with valid JSON only:
{"preference_match": "...", "semantic_distractor": "..."}
\end{verbatim}
}

\paragraph{Rewrite prompt B (Appendix~\ref{app:generator}).}
{\footnotesize
\begin{verbatim}
Below is an opinion someone expressed about a social or political
issue.

Opinion: "{anchor}"

Rewrite it twice:

(a) PREFERENCE MATCH - a statement this person would also agree
    with. It must take the same side of the issue and reflect the
    same underlying values, but share as few words, phrases, and
    grammatical patterns with the original as possible. Change the
    framing entirely.

(b) SEMANTIC DISTRACTOR - a statement this person would reject.
    Reuse the original's words, phrasing, and sentence structure
    almost verbatim, changing only what is needed to reverse the
    position it takes.

For example, given "School uniforms should be mandatory in all
public schools":
(a) "Letting kids dress identically removes distractions and levels
    the social playing field - every district ought to adopt this
    policy."
(b) "School uniforms should not be mandatory in any public school."

Respond with valid JSON only:
{"preference_match": "...", "semantic_distractor": "..."}
\end{verbatim}
}

\paragraph{Rewrite prompt C (Appendix~\ref{app:generator}).}
{\footnotesize
\begin{verbatim}
You are building a benchmark that separates stance understanding
from lexical similarity.

Statement: "{anchor}"

Produce:
1. "semantic_distractor": flip the statement's stance while copying
   its vocabulary and syntax as closely as possible (maximal surface
   overlap, opposite position).
2. "preference_match": keep the statement's stance and values intact
   while expressing them in completely new words and structure
   (minimal surface overlap, same position).

Answer with a single JSON object containing exactly those two keys.
\end{verbatim}
}

\paragraph{Mirrored rewrite for easy controls (Appendix~\ref{app:polarity}).}
{\footnotesize
\begin{verbatim}
You are given a person's opinion statement on a political or social
topic.

Original statement: "{anchor}"

Generate two rewritten versions:

1. SEMANTIC DISTRACTOR (opposite stance, different words):
   - Express the OPPOSITE stance on the same underlying issue
   - Use completely DIFFERENT vocabulary, framing, and sentence
     structure
   - Goal: MINIMIZE word overlap while having opposite meaning
   - Example: "Renewable energy must be prioritized above all else"
     -> "Cheap, reliable fossil fuels are what people actually
     need - chasing wind and solar comes at too great a cost."

2. PREFERENCE MATCH (same stance, same words):
   - Keep nearly IDENTICAL wording and sentence structure as the
     original
   - Preserve the SAME stance and values, changing only a word or two
   - Goal: MAXIMIZE word overlap while keeping the same meaning
   - Example: "Renewable energy must be prioritized above all else"
     -> "Renewable energy must be prioritized above everything else"

Respond with valid JSON only:
{"semantic_distractor": "...", "preference_match": "..."}
\end{verbatim}
}


\end{document}